\documentclass[Afour,sageh,times]{sagej}

\usepackage{float}
\usepackage{moreverb}
\usepackage{amsmath}
\usepackage{epsfig}
\usepackage{multirow}
\usepackage[table]{xcolor}
\usepackage{graphicx}
\usepackage{soul}
\usepackage{mathrsfs}
\usepackage{flushend}
\usepackage{tabularx}
\usepackage{array}
\usepackage{booktabs}
\usepackage{enumitem}
\usepackage{threeparttable}

\definecolor{best}{RGB}{255,230,204}
\definecolor{second}{RGB}{220,235,255}

\usepackage[colorlinks,
            bookmarksopen,
            bookmarksnumbered,
            citecolor=red,
            urlcolor=red,
            linkcolor=red]{hyperref}

\newcommand\BibTeX{{\rmfamily B\kern-.05em \textsc{i\kern-.025em b}\kern-.08em
T\kern-.1667em\lower.7ex\hbox{E}\kern-.125emX}}

\def\volumeyear{2026}

\begin{document}

\setcounter{secnumdepth}{2}

\runninghead{Chen et al.}

\title{OpenSGA: Efficient 3D Scene Graph Alignment in the Open World}

\author{Gang~Chen\affilnum{1}, 
Sebastián~Barbas~Laina\affilnum{2}, Stefan Leutenegger\affilnum{3}, and Javier~Alonso-Mora\affilnum{1}}

\affiliation{\affilnum{1}Autonomous Multi-Robots Lab, Department of Cognitive Robotics, School of Mechanical Engineering,
        Delft University of Technology, 2628 CD, Delft, Netherlands\\
\affilnum{2}Mobile Robotics Lab, School of Computation, Information and Technology, Technical University of Munich\\
\affilnum{3}Mobile Robotics Lab, Department of Mechanical and Process Engineering,  ETH Z\"urich
}

\corrauth{Gang~Chen, g.chen-5@tudelft.nl}

\begin{abstract}
Scene graph alignment establishes object correspondences between two 3D scene graphs constructed from partially overlapping observations. 
This enables efficient scene understanding and object-level relocalization when a robot revisits a place, as well as global map fusion across multiple agents.
Such capabilities are essential for robots that require long-term memory for long-horizon tasks involving interactions with the environment.
Existing approaches mainly focus on subscan-to-subscan (S2S) alignment and depend heavily on geometric point-cloud features, leaving frame-to-scan (F2S) alignment and open-set vision–language features underexplored. 
In addition, existing datasets for scene graph alignment remain small-scale with limited object diversity, constraining systematic training and evaluation.
We present a unified and efficient scene graph alignment framework that predicts object correspondences by fusing vision–language, textual, and geometric features with spatial context. The framework comprises modules such as a distance-gated spatial attention encoder, a minimum-cost-flow-based allocator, and a global scene embedding generator to achieve accurate alignment even under large coordinate discrepancies.
We further introduce ScanNet-SG, a large-scale dataset generated via an automated annotation pipeline with over 700k samples, covering 509 object categories from ScanNet labels and over 3k categories from GPT-4o-based tagging.
Experiments show that our method achieves the best overall performance on both F2S and S2S tasks, substantially outperforming existing scene graph alignment methods. Our code and dataset are released at: 
\url{https://autonomousrobots.nl/paper_websites/opensga}.
\end{abstract}


\maketitle

\section{Introduction} \label{Section: Introduction} 
3D scene graphs (\cite{armeni20193d}) are emerging as a powerful representation for embodied agents operating in open-world environments, as they provide a compact and scalable abstraction that integrates semantic, geometric, and spatial information about objects. 
A typical 3D scene graph models the environment as nodes corresponding to objects and edges encoding relationships between them.
This structured representation has demonstrated increasing value across a wide range of robotics tasks, including open-world navigation (\cite{liu2025delta,dai2024optimal,devarakonda2024orionnav,tang2025openobject}), mobile manipulation (\cite{gu2024conceptgraphs,honerkamp2024language,rana2023sayplan}), simultaneous localization and mapping (SLAM) (\cite{bavle2023s,peterson2025roman,11024207}), and high-level scene understanding and reasoning (\cite{ge2024commonsense,gorlo2024long}).

\begin{figure}
    \centering
    \includegraphics[width=1\linewidth]{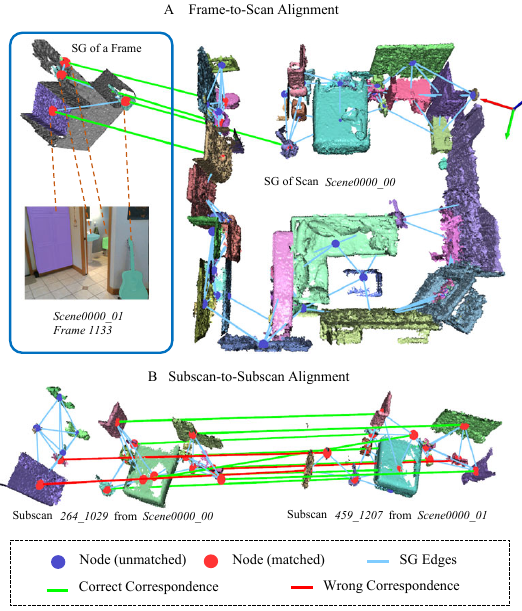}
    \caption{A visualization of the F2S (A) and S2S (B) alignment tasks in our ScanNet-SG dataset. Correspondences are predicted by our released high-performance model, and correct correspondences are illustrated with green lines. SG denotes scene graph. In (A), the scene graph of Frame 1133 in \texttt{Scene0000\_01} is shown on the left, while the scene graph of the full scan (\texttt{Scene0000\_00}) of the same scene is shown on the right. In (B), two subscans constructed from the same scene with an overlap ratio of 0.87 are presented.}
    \label{fig: cover}
\end{figure}

In many of these applications, robots must continuously integrate observations across time and across agents by revisiting previously observed places and maintaining a consistent object-level representation. This requires identifying shared objects across scene graphs of the same environment despite viewpoint changes and partial overlap, which is formalized as the scene graph alignment problem (\cite{sarkar2023sgaligner}) and remains challenging due to viewpoint-dependent appearance variation, partial observations, and the diverse set of objects encountered in real-world environments.

In practice, scene graph alignment arises in two complementary settings: frame-to-scan (F2S) alignment, which aligns a partial observation (e.g., a single RGB-D frame) to a global scan-level scene graph under limited visible context, and subscan-to-subscan (S2S) alignment, which aligns two partial reconstructions (e.g., subscans / submaps) captured from different viewpoints, time periods, or agents. 
F2S is particularly useful for scene understanding and object relocalization when a robot revisits a place or resumes operation after charging or recovery, and must quickly align a partial observation to its global map.
S2S is particularly useful for global map fusion, allowing a single robot or multiple agents to align and merge independently built submaps.

Some recent works (\cite{sarkar2023sgaligner,xie2024sg,11024207,singh2025sgaligner++}) have started to formalize this problem, demonstrating scene graph alignment as a key primitive for single or multiple robot mapping and navigation. 
However, these works remain limited in task coverage, matching strategies, and dataset scale, which constrains systematic evaluation in robotics settings. 

(1) \textit{Tasks}: Although existing methods have been proposed for the S2S alignment, the F2S alignment setting is still underexplored. SceneGraphLoc (\cite{miao2024scenegraphloc}) introduces a cross-modal localization method for the F2S task, but focuses on room-level coarse localization rather than object-level node alignment.  

(2) \textit{Matching methods}: 
Most approaches rely heavily on geometric registration cues from point clouds, which are unreliable under partial observations due to sparse overlap and viewpoint-dependent occlusions. 
Although some methods incorporate semantic embeddings (\cite{11024207,singh2025sgaligner++}), the potential of modern vision–language models (VLMs) derived open-set object features remains underexplored for robust scene graph alignment.

(3) \textit{Datasets}: Current datasets do not provide large-scale, diverse open-set object alignment data. For instance, SGAligner (\cite{sarkar2023sgaligner}) constructs a benchmark with 17k samples on 3D SSG (\cite{wald2020learning}), but the label space remains limited to 160 classes. In contrast, SG-Reg (\cite{11024207}) includes a broader open-set vocabulary but releases only 100 alignment samples.
These limitations motivate a new large-scale dataset and learning-based alignment framework that support open-world objects under partial and cross-view overlap in both S2S and F2S settings.

In this work, we propose an efficient scene graph alignment framework that fuses VLM features from GroundingDINO (\cite{liu2024grounding}), BERT features (\cite{reimers2019sentence}), and 3D bounding box features with spatial context to predict correspondences between two scene graphs. The framework incorporates a distance-gated spatial attention (DGSA) encoder for contextual feature fusion, a matching score predictor with lightweight and high-performance variants, and a minimum-cost-flow (MCF)–based allocator for many-to-one matching. A learnable class embedding is further introduced to facilitate scalable alignment in big multi-scene environments.
To enable large-scale training and evaluation, we also develop an automated annotation pipeline that generates open-set 3D scene graphs from RGB-D images and poses by integrating foundation models with point cloud processing tools. Applying this pipeline to ScanNet (\cite{dai2017scannet}), we construct a dataset supporting both F2S and S2S alignment tasks, in which each object node is enriched with multimodal attributes.
Figure~\ref{fig: cover} presents the dataset and representative F2S and S2S alignment results generated by the high-performance variant of our model.

Our main contributions include:
\begin{itemize}[leftmargin=*, itemsep=2pt, topsep=2pt, labelsep=0.5em]

\item We introduce an end-to-end 3D scene graph alignment framework for object-level correspondence prediction, enabling robust alignment under partial observations and viewpoint changes. Compared with baseline methods, our approach improves accuracy by 6.4\%--13.7\% and F1 score by 4.1\%--11.2\% in different test groups. Relative to the state-of-the-art method that also requires training, our framework reduces both training and inference time by at least 60\% while maintaining superior performance.

\item We propose a distance-gated spatial attention encoder to capture node context under large coordinate discrepancies, an MCF-based allocator for many-to-one association, and a global embedding module for efficient matching across multi-scene environments. Together, these components improve alignment accuracy on both F2S and S2S tasks.

\item We develop an automated annotation pipeline for constructing 3D scene graphs enriched with textual and VLM features. Building on this pipeline, we introduce ScanNet-SG, a large-scale dataset and benchmark that supports both F2S and S2S alignment tasks in indoor environments. ScanNet-SG contains over 700k annotated alignment samples, making it approximately 45× larger than existing datasets, and covers more than 500 object classes in the SG-509 subset and 3k classes in the SG-GPT subset, representing 3× and 18× increases in category diversity, respectively.
\end{itemize}

We also tested using the predicted correspondences to support downstream point cloud registration, demonstrating their practical utility for registering point clouds with low overlap ratios.
We release both lightweight and high-performance models of our method to support different efficiency--accuracy trade-offs.


\section{Related Works} \label{Section: Related Work}
\subsection{3D Scene Graph}
In recent years, 3D scene graphs have emerged as a tool for environment representation, encoding perceptually relevant elements in a graph, where nodes correspond to objects or regions, and edges encode relationships such as spatial or semantic interactions (\cite{armeni20193d}). Each node is often associated with rich semantic and geometric attributes, such as semantic labels, text embeddings, point clouds, 3D bounding boxes, and other node-level properties.
As a compact and scalable environment representation (\cite{gu2025mr,hughes2024foundations}), 3D scene graphs have recently gained significant attention in robotics. They have been used for open-world navigation and mobile manipulation (\cite{gu2024conceptgraphs,honerkamp2024language,rana2023sayplan,liu2025delta,dai2024optimal,devarakonda2024orionnav,tang2025openobject,chen2025synergai}), as well as to improve the accuracy of SLAM (\cite{bavle2023s,peterson2025roman}), infer the location of unseen objects (\cite{ge2024commonsense}), perform visual relocalization (\cite{oliveira2025regrace}), predict long-term human trajectories (\cite{gorlo2024long}), and support scene generation (\cite{liu2025controllable}). These applications highlight the versatility of 3D scene graphs as a unified high-level representation for perception, decision-making, and planning.

Hydra (\cite{Hughes2022Hydra,hughes2024foundations}) enables real-time hierarchical scene graph construction from RGB-D data by integrating SLAM, object detection, and geometric verification, while Hydra-Multi (\cite{chang2023hydra}) extends this capability to collaborative multi-robot mapping through hierarchical graph fusion. With the emergence of foundation models, such as VLMs (\cite{liu2024grounding,radford2021learning,openai_gpt-4o_2024}) and segmentation models such as Segment Anything (\cite{kirillov2023segment,zhao2023fast}), recent works have explored building open-world 3D scene graphs that support open-set object categories (\cite{koch2024open3dsg,chen2024clip,maggio2024clio,zhang2025open,wu2023incremental,strader2024indoor,olivastri2024multi,11024207}). These methods typically use SLAM for localization and dense geometric reconstruction, while VLMs and segmentation models are used to detect object instances, associate geometric points with each node, and infer spatial or semantic relationships between nearby nodes to construct graph edges.
We do not focus on real-time scene graph construction. Instead, we propose an offline pipeline to build scene graphs for generating training and evaluation data for scene graph alignment. 
Our primary focus is on the scene graph alignment described in the following sections.

\subsection{Scene Graph Alignment}
Scene graph alignment aims to establish correspondences between nodes in two scene graphs that represent partially overlapping observations of the same environment (\cite{sarkar2023sgaligner}). 
Unlike geometric registration, which focuses on estimating spatial transformations, scene graph alignment operates at the object level by matching nodes based on semantic, geometric, and relational attributes.
Accurate alignment is critical for robotics applications, such as localization (\cite{bavle2023s,peterson2025roman}) or relocalization (\cite{oliveira2025regrace}), long-term or multi-gent mapping (\cite{11024207}), and downstream point cloud registration (\cite{sarkar2023sgaligner,xie2024sg}), where reliable object correspondence matters.

SGAligner (\cite{sarkar2023sgaligner}) first formalized scene graph alignment by encoding object point clouds as node features and uses graph attention networks to aggregate spatial and structural context from neighboring nodes, followed by contrastive learning to predict correspondences.
SG-PGM (\cite{xie2024sg}) further improves alignment by fusing the semantic point clouds through joint node-level and fragment-level encoders, enabling more robust matching under partial overlap.
More recent works incorporate additional modalities to improve robustness and generalization.
SG-Reg (\cite{11024207}) integrates point cloud feature, Bert embedding of the object label, and bounding box features for efficient and generalizable alignment, while SGAligner++ (\cite{singh2025sgaligner++}) further incorporates CAD features, constructing unified scene graphs that preserve structural relationships across observations. 
In addition to learned matching methods, ROMAN (\cite{peterson2025roman}) performs zero-shot scene graph alignment by leveraging CLIP (\cite{radford2021learning}) embeddings, PCA-based geometric features extracted from object point clouds, and a unified graph-theoretic global data association framework. 

Despite these advances, most existing scene graph alignment methods rely predominantly on geometric features derived from object point clouds, which are sensitive to incomplete observations and sparse overlap. While VLM features have been exploited in zero-shot approaches such as ROMAN (\cite{peterson2025roman}), their incorporation into trainable matching architectures remains largely underexplored.
Furthermore, prior works have primarily focused on the S2S alignment task, where both scene graphs typically have comparable spatial coverage. The F2S alignment task, in which one graph represents only a small partial observation of another much larger scene graph, remains insufficiently studied. In this work, we address both the F2S and S2S alignment tasks within a unified framework, evaluated on different subsets of the proposed dataset.


\subsection{Dataset}
Several datasets with 3D scene graph annotations have been proposed in recent years, such as 3D SSG (\cite{wald2020learning}), SG-FRONT (\cite{ge2024commonsense}), VLA-3D (\cite{zhang2024vla}), and SG3D (\cite{sg3d}). These datasets construct scene graphs on top of semantically annotated point cloud scans for navigation and mobile manipulation tasks. However, they do not contain scene graph pairs constructed from partially overlapping observations of the same environment, and thus do not support the scene graph alignment task.

Few datasets explicitly supporting 3D scene graph alignment have been proposed. Building upon the scene graph annotations of 3DSSG, SGAligner (\cite{sarkar2023sgaligner}) introduces a dataset for aligning partially overlapping 3D scene graphs. In this dataset, subscans are generated from globally aligned scans in 3RScan (\cite{wald2019rio}) with controlled overlap ratios. 
Each subscan is converted into a sub-scene graph, and nodes within overlapping regions are treated as ground-truth correspondences.
The dataset contains approximately 15.1k training pairs and 1.9k validation pairs, covering around 160 object classes. 
Later, SG-Reg (\cite{11024207}) proposed a smaller dataset of approximately 4k samples tailored specifically for S2S alignment.

Although both SGAligner and SG-Reg provide object-level point clouds and semantic labels, they do not include vision–language features, which we identify as a critical cue for robust open-world scene graph matching. Moreover, these datasets focus exclusively on S2S alignment. 
The F2S setting, where a partial observation must be matched to a larger global map, remains unexplored. 
In contrast, our dataset is substantially larger, covering over 500 and 3k object classes across different groups and containing more than 700k samples. It additionally incorporates GroundingDino (\cite{liu2024grounding}) vision–language features for each node and supports both S2S alignment and F2S matching, enabling evaluation in more diverse open-world scenarios.

\section{Problem Definition} \label{Section: Problem Definition}
OpenSGA is mainly targeted to solve the open-world 3D semantic scene graph alignment problem.
A 3D scene graph is denoted as $\mathcal{G}=\{\mathcal{V}, \mathcal{E}\}$, where $\mathcal{V}$ is a set of nodes that represent real objects in the environment and $\mathcal{E}$ is a set of edges that connect the nodes. 

(a) \textit{Nodes}:
Each node $\boldsymbol{v}_i \in \mathcal{V}$ corresponds to a real-world object and contains multimodal attributes, such as visual--language features $\boldsymbol{f}_{\mathrm{vl}}$, text embedding $\boldsymbol{f}_{\mathrm{t}}$, and a geometric feature $\boldsymbol{f}_{\mathrm{g}}$. 
Let $\boldsymbol{x}_i \in \mathbb{R}^3$ denote the object’s 3D position.  
The node representation is then
\begin{equation}
    \boldsymbol{v}_i
    =
    [\boldsymbol{x}_i,\;
    \boldsymbol{f}_{\mathrm{vl},i},\;
    \boldsymbol{f}_{\mathrm{t},i},\;
    \boldsymbol{f}_{\mathrm{g},i}],
\end{equation}
where $\boldsymbol{f}_{\mathrm{g},i}$ in our method is the size of the object's 3D oriented bounding box, while point-based geometric descriptors used in prior works (\cite{11024207,xie2024sg}) are included only for comparison.
The text embeddings $\boldsymbol{f}_{\mathrm{t}}$ and the vision–language features $\boldsymbol{f}_\mathrm{vl}$ are obtained by encoding the label text with SBERT (\cite{reimers2019sentence}) and by extracting the hidden-layer representation of GroundingDINO (\cite{liu2024grounding}) during object detection, respectively. Details of these features are provided in Section~\ref{Section: Scene Graph Building and Dataset Construction}.

(b) \textit{Edges}:
An undirected edge $\boldsymbol{e}_{ij}\in\mathcal{E}$ connects nodes $\boldsymbol{v}_i$ and $\boldsymbol{v}_j$ and stores their Euclidean distance,
\begin{equation}
    d_{ij} = \|\boldsymbol{x}_i - \boldsymbol{x}_j\|.
\end{equation}
To avoid fully connected graphs, we connect only the $N$ nearest neighbors for each node.

(c) \textit{Alignment Problem}:
Given two partially overlapped 3D scene graphs $\mathcal{G}^A$ and $\mathcal{G}^B$ captured from the same environment, the goal is to find the correct node correspondences
\begin{equation}
    \mathcal{M}^{AB}
    = 
    \{(\boldsymbol{v}^A_i,\boldsymbol{v}^B_j)
    \mid
    \boldsymbol{v}^A_i\in\mathcal{G}^A,\;
    \boldsymbol{v}^B_j\in\mathcal{G}^B\}.
\end{equation}
Only nodes that physically correspond to the same real object should be matched. 
Note some nodes have no matches because the graphs are often built from different views and may cover different areas of the scene. 

In practical robotic navigation scenarios, we assume that $\mathcal{G}^A$ and $\mathcal{G}^B$ originate from different sources in the two tasks described in Section~\ref{Section: Introduction}.
In the F2S task, $\mathcal{G}^A$ is constructed from the current observation, while $\mathcal{G}^B$ is derived from a more complete map accumulated from previous observations. The node positions in $\mathcal{G}^B$ are represented in the global reference frame of the previously built map. In contrast, the node positions in $\mathcal{G}^A$ are defined in the current camera frame, whose pose may be arbitrary and unrelated to the global reference frame of $\mathcal{G}^B$. This design decouples object alignment from ego-pose estimation, ensuring that correspondences are established based on object features and spatial relationships rather than similarities in global poses.
In the S2S task, both $\mathcal{G}^A$ and $\mathcal{G}^B$ are constructed from sequences of images that capture the same scene with partial overlap. 
These graphs represent submaps generated either by multiple robots or by a single robot at different times. 
We assume a static environment for both tasks.
Since both tasks are inherently node-level alignment problems, we address them using a unified network architecture.
The similarities and differences of the two tasks will be further discussed in Section~\ref{Section: Scene Graph Building and Dataset Construction}.

\section{Scene Graph Alignment} \label{Section: Scene Graph Matching}
In this section, we first present our network designed for F2S and S2S alignment in \ref{section: alignment network}. We assume that the graphs to align have been built and represented as $\mathcal{G}^A$ and $\mathcal{G}^B$ (specifications of how the scene graphs are built can be found later in Section~\ref{Section: Scene Graph Building}). 
As a large-scale environment can contain multiple scenes to align with, we introduce a global graph retrieval module that computes discriminative graph-level embeddings in \ref{section: global embedding}. 
The loss functions used in training are presented in \ref{section: loss function}.

\subsection{Alignment Network} \label{section: alignment network}
Our alignment network is designed to find the match $\mathcal{M}^{AB}$ given input 3D scene graphs $\mathcal{G}^{A}$ and $\mathcal{G}^B$. The architecture of the network is shown in Fig.~\ref{fig:system}. We divide our network into three parts: encoder, matcher and allocator.
Each part is introduced in detail in the following:

\begin{figure*}[h]
\centering
\includegraphics[width=1\linewidth]{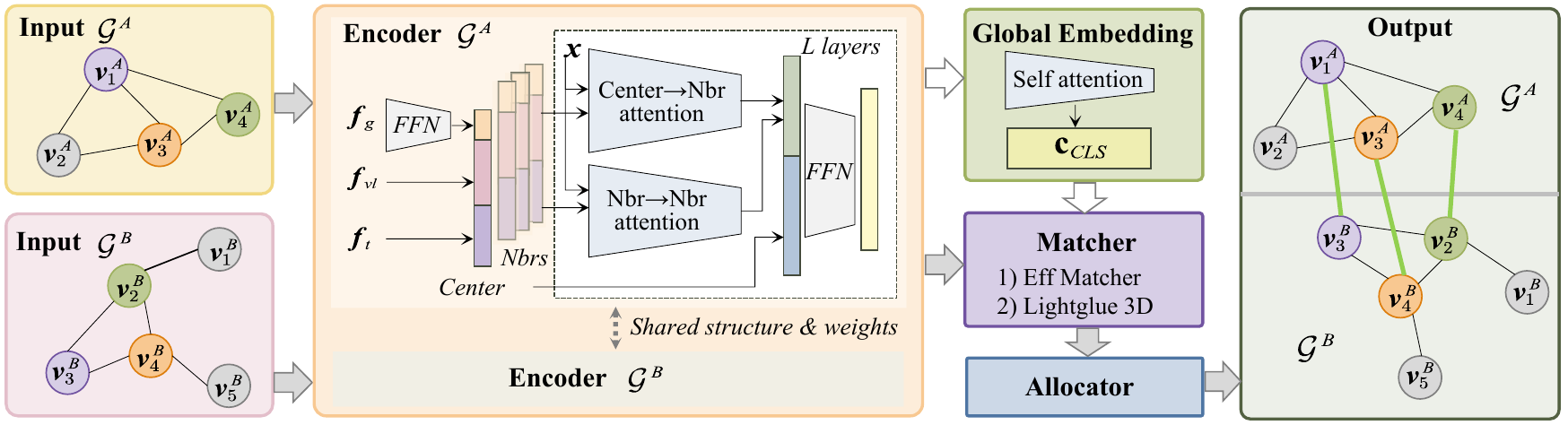}
\caption{The system architecture of our 3D scene graph alignment network. From left to right: input scene graphs, alignment network composed of encoder, matcher and allocator, output matching result. Each node is composed of 3D position $\boldsymbol{x}$, visual language feature $\boldsymbol{f}_\mathrm{vl}$, text feature $\boldsymbol{f}_t$ and geometry feature $\boldsymbol{f}_g$. The green lines in the output block illustrate the matched nodes. ``Nbr'' is the abbreviation of ``neighbor''. The global embedding module is connected with white arrows, as it is used only in large environments containing multiple scenes.}
\label{fig:system}
\end{figure*}

\subsubsection{\textbf{Encoder}}
The encoder aims to generate a feature vector that encodes the essential characteristics of each node as well as its surrounding neighborhood under open-world variations. 
The encoders for $\mathcal{G}^{A}$ and $\mathcal{G}^B$ share the same structure and weights. For clarity, we omit the superscripts $A$ and $B$ in the following description.

The text embeddings $\boldsymbol{f}_{\mathrm{t}}$ and the vision–language features $\boldsymbol{f}_{\mathrm{vl}}$ are already high-dimensional embeddings produced by pre-trained models (\cite{reimers2019sentence,liu2024grounding}).
In contrast, the geometric descriptor $\boldsymbol{f}_{\mathrm{g}}$ is a normalized 3D vector representing the size of the object’s oriented bounding box. We first encode $\boldsymbol{f}_{\mathrm{g}}$ using a feed-forward network (FFN) with one hidden layer to obtain a higher-dimensional embedding.
By concatenating this embedding with $\boldsymbol{f}_\mathrm{vl}$ and $\boldsymbol{f}_{\mathrm{t}}$, we obtain the initial node feature vector, illustrated as a three-block stacked column in the encoder portion of Fig.~\ref{fig:system} and denoted as $\boldsymbol{c}_i$ hereafter.

While this initial embedding captures only the intrinsic properties of a node (center node), it does not incorporate any context or neighborhood information.
Prior works have demonstrated that local context is critical for accurate scene graph alignment (\cite{sarkar2023sgaligner,11024207,xie2024sg}). However, it is still nontrivial to get a good embedding since it needs to be inherently translation and rotation invariant to ensure consistent representations even when the two input graphs $\mathcal{G}^{A}$ and $\mathcal{G}^{B}$ are defined in very different coordinate frames.

The current state-of-the-art, Triplet-boosted Graph Neural Network (T-GNN) (\cite{xie2024sg}), addresses this issue by constructing triplet features that incorporate the edge length and angle of two randomly selected neighbor nodes and the center node. 
However, T-GNN introduces two key limitations:
(i) it enforces only yaw-rotation invariance, while pitch and roll invariance are not supported since the method relies on defining an anti-clockwise ordering of neighbors for constructing triplet features;
(ii) at least two neighbor nodes are required to make a Triplet thus for a frame with only one neighbor node, the neighbor information, which is still useful, is skipped.

Our method begins by leveraging the inherent global rigidity of local 3D scene graphs.
Under the assumption that each node has at least three neighbors in non-coplanar positions, the center node and its neighbors form a globally rigid subgraph, whose structure is uniquely determined by the set of pairwise distances between nodes.
Therefore, the geometric layout of the objects can be fully described using only the distances, which naturally yields translation- and rotation-invariant representations.
Building upon this insight, our method encodes neighborhood geometry exclusively through center-to-neighbor distances and neighbor-to-neighbor distances. 
Specifically, we make two distance-gated attention blocks:

(a) \textit{Center to neighbor attention}:
For each center node $\boldsymbol{v}_i$, we first calculate the distance $d_{ij}$ to each neighbor node $\boldsymbol{v}_j$. 
The distance is encoded by a sinusoidal positional encoder described in (\cite{vaswani2017attention}). Let $\text{PE}(d_{ij}), \ i \neq j$ denote the encoded distance embedding. The neighbor feature of $\boldsymbol{v}_j$ is defined as
\begin{equation}
    \boldsymbol{h}_{ij} = \left( \text{PE}(d_{ij}) \; \Vert \; \boldsymbol{c}_j \right)
\end{equation}

We then compute the attention from the center node to its neighbors by:
\begin{equation}
    \boldsymbol{q}_i = \boldsymbol{W}_\mathrm{q} \boldsymbol{c}_i,\quad
    \boldsymbol{k}_{ij} = \boldsymbol{W}_\mathrm{k} \boldsymbol{h}_{ij},\quad
    \boldsymbol{v}_{ij} = \boldsymbol{W}_\mathrm{v} \boldsymbol{h}_{ij} 
\end{equation}
\begin{equation}
    \alpha^{raw}_{ij} = 
       \frac{\boldsymbol{q}_i^\top \boldsymbol{k}_{ij}}
            {\sqrt{d_h}}
\end{equation}
where $\boldsymbol{q}_i$, $\boldsymbol{k}_{ij}$ and $\boldsymbol{v}_{ij}$ are the query, key and value while $\boldsymbol{W}_\mathrm{q}$, $\boldsymbol{W}_\mathrm{k}$ and $\boldsymbol{W}_\mathrm{v}$ are the corresponding learnable weight matrices. $d_h$ is the dimensionality of each attention head. $\alpha^{raw}_{ij}$ is the calculated raw attention score and is further gated by:
\begin{equation}
    \alpha_{ij}
    =
    \mathrm{g}(d_{ij})\cdot \alpha^{raw}_{ij},
\end{equation}
where $\mathrm{g}(d_{ij})$ denotes the distance gate function implemented by a lightweight gating network. The network consists of a two-layer MLP followed by a sigmoid activation, producing a scalar value in $(0,1)$. 
This distance gate explicitly adds geometric prior into the attention to enforce biased attention based on distance, e.g., close neighbors should get more attention.

Then the attention weight is normalized with softmax and the output $\boldsymbol{o}_i^{(\mathrm{c} \rightarrow \mathrm{n})}$ is calculated with
\begin{equation}
    \boldsymbol{o}_i^{(\mathrm{c} \rightarrow \mathrm{n})} = \sum_{j \in \mathcal{N}(i)} \operatorname{softmax}_j(\alpha_{ij})\boldsymbol{v}_{ij}.
\end{equation}
where the superscript $(c\rightarrow n)$ stands for center to neighbors and $\mathcal{N}(i)$ is the neighbor ID set of the $i$th node.

(b) \textit{Neighbor to neighbor attention}:
The neighbor to neighbor attention of the $i$th node works similarly as the center to neighbor attention and can be formulated as the following:

\begin{equation}
    \boldsymbol{q}_{ij}=\boldsymbol{W}'_\mathrm{q}\boldsymbol{h}_{ij},\;
    \boldsymbol{k}_{ik}=\boldsymbol{W}'_\mathrm{k}\boldsymbol{h}_{ik},\;
    \boldsymbol{v}_{ik}=\boldsymbol{W}'_\mathrm{v}\boldsymbol{h}_{ik}, 
\end{equation}
\begin{equation}
    \beta_{jk} = \mathrm{g}(d_{jk})
    \frac{\boldsymbol{q}_{ij}^\top \boldsymbol{k}_{ik}}
         {\sqrt{d_h}},
\end{equation}
\begin{equation} \label{Eq: n2n output}
    \boldsymbol{o}_i^{(\mathrm{n} \rightarrow \mathrm{n})} 
    = \frac{1}{|\mathcal{N}(i)|}
         \sum_{j,k\in\mathcal{N}(i), j \neq k}
         \operatorname{softmax}_k(\beta_{jk})\boldsymbol{v}_{ik}.
\end{equation}
where $i\neq j \neq k$ and  $\boldsymbol{W}'_\mathrm{q}$, $\boldsymbol{W}'_\mathrm{k}$ and $\boldsymbol{W}'_\mathrm{v}$ are learnable weight matrices. The output $\boldsymbol{o}_i^{(\mathrm{n} \rightarrow \mathrm{n})}$ in (\ref{Eq: n2n output}) is the average pooling over all neighbors. 

Note the computational complexity of center to neighbor attention is $O(N)$ while that of neighbor to neighbor attention is $O(N^2)$. In practice, we set the maximum neighbor number $N=4$ to ensure efficiency.

After $\boldsymbol{o}_i^{(\mathrm{c} \rightarrow \mathrm{n})}$ and $\boldsymbol{o}_i^{(\mathrm{n} \rightarrow \mathrm{n})}$ are computed, we get the combined neighbor feature embedding using residual (\cite{he2016deep}) and LayerNorm (\cite{ba2016layer}) as follows:
\begin{equation}
    \tilde{\boldsymbol{c}}_i
    = 
    \operatorname{LayerNorm}\!\left(
        \boldsymbol{c}_i
        + \boldsymbol{W}_\mathrm{o}(\boldsymbol{o}_i^{(\mathrm{c} \rightarrow \mathrm{n})}+\boldsymbol{o}_i^{(\mathrm{n} \rightarrow \mathrm{n})})
    \right)
\end{equation}
where $\boldsymbol{W}_\mathrm{o}$ stands for the weight of a single MLP layer.

The final node representation for alignment is obtained by concatenating $\tilde{\boldsymbol{c}_i}$ and $\boldsymbol{c}_i$. The concatenated feature is then passed through a projection feed-forward network (FFN) to match the input dimensionality required by the matcher.
When the number of neighbors is fewer than three, the local 3D scene graph composed of the ego node and its neighbors is no longer globally rigid. Nevertheless, the distance-gated attention still serves as an effective cue for modeling neighborhood relationships, and our method performs well in practice.

\subsubsection{\textbf{Matcher}}
Given the encoded node sets from $\mathcal{G}^A$ and $\mathcal{G}^B$, this module predicts matching scores, i.e., the pairwise similarities between nodes in $\mathcal{G}^A$ and $\mathcal{G}^B$. For simplicity, we refer to this matching score prediction module as the matcher in the following sections.

Our framework provides two variants of the matcher.
The first is a lightweight cosine-similarity-based matcher. It computes pairwise similarities using normalized dot products. To handle unmatched nodes, we incorporate a learnable ``dustbin'' row and column that provide the model with an explicit option for non-matches.

The second variant adapts LightGlue (\cite{lindenberger2023lightglue}), a popular and efficient 2D visual keypoint matching network, to the 3D setting in order to estimate similarity scores. Specifically, we extend keypoints to 3D, apply 3D coordinate normalization, and lift the positional encoding to 3D accordingly.
The input feature dimension is also adjusted to be compatible with our model. 
The adapted version is referred to as LightGlue3D in Fig.~\ref{fig:system}. Compared to the first matcher, LightGlue3D achieves a higher performance but requires more GPU memory as well as longer training and inference time.

\subsubsection{\textbf{Allocator}}
Our matcher outputs a similarity score matrix. Let $I$ and $J$ denote the number of nodes in $\mathcal{G}^A$ and  $\mathcal{G}^B$, respectively. Suppose $\boldsymbol{P} \in \left[0, 1 \right]^{I \times J}$ is the similarity score matrix. 
Once $\boldsymbol{P} \in \left[0, 1 \right]^{I \times J}$ is predicted, the final matching result $\mathcal{M}^{AB}$ needs to be allocated. Note this step is not included in training but uses the algorithm described in the following. 

One commonly used allocation strategy is mutual nearest neighbor (MNN), which we adopt as one of the allocation algorithms in our framework. However, MNN only supports one-to-one matching.
In practice, under-segmentation occasionally occurs (3–5\% of nodes in our dataset). 
For example, two spatially adjacent sofas or connected cabinets may be segmented as a single instance in some frames and ultimately merged into one object in the map, while in the current frame, they can be segmented as two separate instances.
As a result, multiple instances in the frame may correspond to the same object in the map. We further illustrate this issue in Fig.~\ref{fig: many-to-one}.
Therefore, the matching can be many-to-one rather than strictly one-to-one. To address this, we formulate the matching problem as an iterative minimum-cost flow (MCF) problem and introduce a geometric penalty to encourage shape consistency between $\mathcal{G}^A$ and $\mathcal{G}^B$.

We define a virtual source node $\boldsymbol{s}$ and a sink node $\boldsymbol{t}$ and construct the flow in the following.
Each node $\boldsymbol{v}^A_i$ receives one unit of flow from the source:
\begin{equation}
    (\boldsymbol{s} \rightarrow \boldsymbol{v}^A_i), 
    \qquad 
    \mathrm{cap}=1,\quad \mathrm{cost}=0
\end{equation}
where $\mathrm{cap}$ is the capacity. 
For each pair with similarity above a threshold $\tau$ and among the top K candidates in $P$, we add a candidate matching edge, which is
\begin{equation}
    (\boldsymbol{v}^A_i \rightarrow \boldsymbol{v}^B_j), (i,j)\in \mathcal{C},
    \quad
    \mathrm{cap}=1,\quad
    \mathrm{cost}=c_{ij}
\end{equation}
\begin{equation}
    \mathcal{C}
    =
    \left\{
        (i,j)
        \;\middle|\;
        P[i,j] \ge \tau,
        \; j \in \mathrm{Top}\text{-}K(i)
    \right\}
\end{equation}
where $\mathrm{Top}\text{-}K(i)$ denotes the indices of the $K$ largest entries in row $i$ of $P$.
The cost is defined as follows:
\begin{equation}
    c_{ij}^{(t)} = -\log(P[i,j]) + \lambda \mathrm{pen}^{(t-1)}(i,j)
\end{equation}
where $(t)$ represents the current iteration. 
For a candidate correspondence $(i,j)$ and a set of previously selected
matches $\mathcal{M}^{(t-1)} = \{(k,l)\}$, 
we define the geometry penalty as
\begin{equation}
\label{eq:geom_penalty}
    \mathrm{pen}^{(t-1)}(i,j)
    =
    \max_{(k,l)\in\mathcal{M}^{(t-1)}}
    \left|  d^A_{i,k} - d^B_{j,l}    \right|
\end{equation}
which is, pairwise structural consistency term, penalizing matchings that distort relative distances using the matching result from last iteration. 
Here, $d^A_{i,k}$ denotes the Euclidean distance between nodes $i$ and $k$ in $\mathcal{G}^A$, while $d^B_{j,l}$ denotes the distance between nodes $j$ and $l$ in $\mathcal{G}^B$.

Each node in $\mathcal{G}^A$ may also be left unmatched with a constant cost $\text{C}_{\text{unmatched}}$:
\begin{equation}
    (\boldsymbol{v}^A_i \rightarrow \boldsymbol{t}),
    \qquad
    \mathrm{cap}=1,\quad
    \mathrm{cost}=\text{C}_{\text{unmatched}}
\end{equation}

Nodes in $\mathcal{G}^B$ have matching capacity more than one:
\begin{equation}
    (\boldsymbol{v}^B_j \rightarrow \boldsymbol{t}),
    \qquad
    \mathrm{cap}=\text{Cap}_{max},\quad
    \mathrm{cost}=0
\end{equation}
allowing at most $\text{Cap}_{max}$ matches to be assigned to $\boldsymbol{v}^B_j$.

Let $F_{ij}$ denote the binary flow variable on the edge 
$(\boldsymbol{v}^A_i \rightarrow \boldsymbol{v}^B_j)$ 
for $(i,j) \in \mathcal{C}$.
We solve the following discrete optimization problem:
\begin{align}
    \min_{F} \quad 
    & \sum_{(i,j) \in \mathcal{C}} F_{ij}\, c_{ij}^{(t)} \\[4pt]
    \text{s.t.} \quad
    & F_{ij} \in \{0,1\},
      && \forall (i,j) \in \mathcal{C} \label{Eq: constraint1} \\[4pt] 
    & \sum_{j : (i,j) \in \mathcal{C}} F_{ij}
      \;+\;
      F_{i \rightarrow t}
      = 1,
      && \forall i = 1,\dots,I \label{Eq: constraint2}\\[4pt]
    & \sum_{i : (i,j) \in \mathcal{C}} F_{ij}
      \le \text{Cap}_{max},
      && \forall j = 1,\dots,J \label{Eq: constraint3}
\end{align}
where (\ref{Eq: constraint1}), (\ref{Eq: constraint2}), 
and (\ref{Eq: constraint3}) enforce the binary flow, flow conservation, and capacity constraints.

The problem is solved by iteratively solving a linear min-cost flow problem with NetworkX (\cite{hagberg2008exploring}) in CPU and recalculating $\mathrm{pen}^{(t-1)}(i,j)$ until the result has converged or a maximum iteration is reached.
Detailed parameter values are provided in Appendix~\ref{App: Model Parameters}.

Note that in our setting, $\mathcal{G}^B$ is typically constructed from a more complete scan, where spatially adjacent objects with the same semantic label are fused into a single instance, as discussed at the beginning of this subsection. As a result, many-to-one matching can occur in our setting, whereas one-to-many matching does not. Nevertheless, if one-to-many matching were to be considered, the capacity of the edge $(\boldsymbol{s} \rightarrow \boldsymbol{v}_i)$ could be increased to a value greater than one.

\subsection{Global scene embeddings} \label{section: global embedding}

Our alignment network performs node-level matching between two scene graphs. However, in large-scale environments containing multiple scenes, a robot must first identify the most relevant scene corresponding to the current observation before performing detailed node-level matching. Exhaustively matching the current observation against all candidate scenes is computationally inefficient.
To address this, we introduce a global graph retrieval module that computes discriminative graph-level embeddings and retrieves the Top-K candidate scenes for subsequent node-level matching.

Specifically, we incorporate a learnable class embedding $\mathbf{c}_\mathrm{CLS}$ into the encoder. The class token aggregates the graph information via an independent two-layered multi-head self-attention module.
This descriptor is then used to query the most relevant scene graphs from a database containing all candidate scenes, thereby enabling a more efficient matching process.

To select the Top-K graphs for node-level matching, we first perform database filtering based on the cosine similarity between global descriptors. We then apply a weighted reranking step to the retrieved Top-K candidates, combining global similarity with node-level matching confidence. The reranking score is defined as:
\begin{equation}
\label{eq:reranking}
s_{tq} = {\mathbf{c}^q_\mathrm{CLS}}^T \mathbf{c}^t_\mathrm{CLS} \sum_{i, j \in \mathcal{M}} \mathbf{P}[i, j].
\end{equation}

In this setting, $s_{tq}$ is the similarity between a graph query $q$ and a target $t$. The elements $i, j \in \mathcal{M}$ are the set of matched nodes between graphs and $\mathbf{P}$ is the matrix that contains all the scores of the matching elements between the two graphs $q$ and $t$. When doing a matching against a database of graphs, the selected graph to match against is the graph $t$ that yields the highest $s_{tq}$ with the query graph $q$.

\subsection{Loss function} \label{section: loss function}
For node-level correspondence prediction in~\ref{section: alignment network}, methods using the cosine-similarity-based efficient matcher (marked with ``L'' in the results) are trained with the bidirectional InfoNCE loss proposed in CLIP (\cite{radford2021learning}). 
This objective maximizes similarity between matched pairs while minimizing similarity between unmatched pairs, encouraging discriminative embeddings for different objects. 
Methods using lifted LightGlue (\cite{lindenberger2023lightglue}) as the matcher (marked with ``H'' in the results) are trained using the negative log-likelihood (NLL) loss implemented in the LightGlue codebase.

To train the global scene embedding in~\ref{section: global embedding}, we rely on a triplet margin contrastive loss with negative hard-mining. Given a processed batch, we select the positive pair and the hardest negative scan, the one that is closest in embedding space to $\mathbf{c}^a_\mathrm{CLS}$. To train our samples, we use the following loss function
\begin{equation}
d_{ap} =
\left\lVert \mathbf{c}^{a}_{\mathrm{CLS}} 
- \mathbf{c}^{p}_{\mathrm{CLS}} \right\rVert_2,\quad
d_{an} =
\left\lVert \mathbf{c}^{a}_{\mathrm{CLS}} 
- \mathbf{c}^{n}_{\mathrm{CLS}} \right\rVert_2
\end{equation}
\begin{equation}
\mathcal{L}_{\mathrm{triplet}}
= \max \left( d_{ap} - d_{an} + m,\ 0 \right)
\end{equation}
where $\mathbf{c}^a_\mathrm{CLS}$ is the global embedding associated to the graph from the image, $\mathbf{c}^p_\mathrm{CLS}$ is the embedding of the associated scan graph and $\mathbf{c}^n_\mathrm{CLS}$ is the embedding of the mined graph.

\section{Scene Graph Building and Dataset Construction}
\label{Section: Scene Graph Building and Dataset Construction}
This section first introduces our pipeline used to construct 3D scene graphs with the object features described in Section~\ref{Section: Problem Definition}. We apply this pipeline to the ScanNet (\cite{dai2017scannet}) dataset to build ScanNet-SG, our dataset designed for scene graph alignment in both the F2S and S2S tasks. We then present the dataset construction and grouping strategies for the F2S and S2S tasks, respectively.

\subsection{Scene Graph Building} \label{Section: Scene Graph Building}
Our 3D semantic scene graph is constructed from RGB-D images with given camera poses. The overall process consists of five steps:

\subsubsection{1) Image tagging} 
Given an RGB image, we first use either the human annotations provided in ScanNet or an object-tagging model, such as ChatGPT-4o (\cite{openai_gpt-4o_2024}), to generate labels or textual descriptions of the objects present in the image.


\subsubsection{2) Segmentation and VLM feature extraction} The generated text descriptions are provided as prompts to Grounded-SAM (\cite{ren2024grounded}) to obtain instance-level object masks. At the same time, we acquire Visual--language embeddings $\boldsymbol{f}_\mathrm{vl}$ by extracting a 256-dimensional embedding vector from the hidden layers of GroundingDINO (\cite{liu2024grounding}) used within Grounded-SAM. 

\subsubsection{3) Text embedding} Text embeddings $\boldsymbol{f}_{\mathrm{t}}$ are extracted from the object descriptions using SBERT (\cite{reimers2019sentence}), resulting in 384-dimensional embedding vectors.

\subsubsection{4) Bounding box computation} Using the depth image and instance masks, we compute instance-aware point clouds. Oriented 3D bounding boxes are then estimated using PCA-based OBB fitting (\cite{gottschalk1996obbtree,ericson2004real}). The geometric feature $\boldsymbol{f}_{\mathrm{g}}$ corresponds to the normalized size of the bounding box, where normalization is performed with respect to the maximum extent of the entire point cloud. The center of each bounding box is used as the spatial position of the corresponding object.

For a 3D scene graph constructed from a single frame, Steps~1)–4) yield the object nodes. Edges are established by computing pairwise distances between nodes and connecting each node to its $N$ nearest neighbors within a distance threshold $d_{th}$. For scene graphs constructed from scans or subscans comprising multiple frames, the following additional step is applied:

\subsubsection{5) Multiview fusion} We use instance masks and depth images from multiple views to generate per-instance point clouds. Instances corresponding to the same object are merged based on 3D mIoU and $\boldsymbol{f}_{\mathrm{t}}$ similarity. During merging, the visual--language embeddings $\boldsymbol{f}_\mathrm{vl}$ and text embeddings $\boldsymbol{f}_{\mathrm{t}}$ are averaged, while the point clouds from different views are aggregated and downsampled using a voxel grid filter. The bounding box is then recomputed from the merged point cloud to obtain $\boldsymbol{f}_{\mathrm{g}}$.
We further present the details of the multiview fusion step in Appendix~\ref{App: Merging}.

\begin{figure}
    \centering
    \includegraphics[width=1\linewidth]{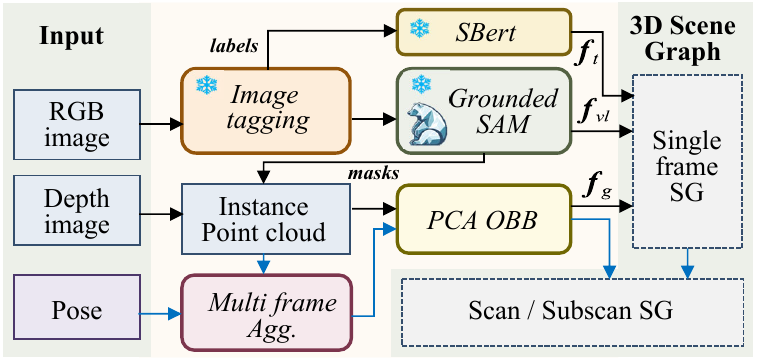}
    \caption{Illustration of our object feature extraction and scene graph construction pipeline. Blue arrows indicate additional steps applied when building scene graphs from multiple frames, such as scans or subscans.}
    \label{fig:graph build}
\end{figure}

The key procedures and data flow is shown in figure~\ref{fig:graph build}.  
In the following, we assume that scene graphs $\mathcal{G}^{A}$ and $\mathcal{G}^{B}$ are constructed using the above procedure. In the frame-to-scan node matching task, $\mathcal{G}^{A}$ represents a smaller graph built from a single frame, while $\mathcal{G}^{B}$ corresponds to a complete scan of the scene. In the subscan-to-subscan node matching task, both $\mathcal{G}^{A}$ and $\mathcal{G}^{B}$ are constructed from incomplete scans of the same scene and share partial overlap.

Our dataset is built upon ScanNet (\cite{dai2017scannet}), which contains over 1{,}500 RGB-D scans spanning 807 indoor scenes. Most scenes contain more than one scan captured from different camera trajectories.
In the following, we present the subsets for frame-to-scan task and subscan to subscan task, respectively.

\subsection{Frame-to-Scan (F2S) Matching} \label{section: F2S}
To support frame-to-scan matching, we generate two groups of frame–scan pairs.
The \textit{Self Scan} group refers to the setting where the frame-level and the scan-level scene graphs originate from the same scan. This setting evaluates whether a robot revisiting a previously observed pose, with access to only a single-frame partial observation, can correctly associate objects with their more complete counterparts in the fused map.
\textit{Cross Scan} Group considers frames sampled from a different scan of the same scene. This setting introduces novel viewpoints and newly appearing and missing regions, thereby reflecting more realistic scenarios.

In the original ScanNet dataset, different scans of the same scene are represented in separate global coordinate frames, and object instances are not associated across scans.
To enable cross-scan matching, we register the scan-level point clouds from different scans of ScanNet using RANSAC-based global registration followed by ICP refinement (\cite{fischler1981random,besl1992method}) to estimate the transformation between scans. To ensure estimation quality, we manually inspect all registration results and discard scan pairs with noticeable misalignment.
We report the estimation performance in Appendix~\ref{App: OpenSGA Dataset}.
Each instance within a scan is assigned a unique identifier in accordance with the labels provided in ScanNet.
Cross-scan instance association within the same scene is then performed using 3D mIoU together with semantic label consistency, as detailed in Step~5) and Appendix~\ref{App: Merging}.

We use Scenes~0–599 (1{,}276 scans) to construct the training set and Scenes~600–705 (237 scans) for testing. Frames are sampled every three frames from the original sequences. In total, the dataset contains approximately 557.0k training pairs and 87.9k test pairs, where 48.3\% are generated from \textit{Self Scan} and 52.7\% from \textit{Cross Scan}. The objects in this dataset span 509 unique semantic labels, and therefore we refer to this benchmark as \textit{ScanNet-SG-509}.

To further increase object diversity and improve open-world generalization, we additionally use GPT-4o-mini (\cite{openai_gpt-4o_2024}) for image-based object tagging. 
Recent study \cite{xie2026multimodal} has shown that tags generated with large VLMs achieve strong agreement with human labels and can support downstream learning tasks effectively.
Compared to the original ScanNet annotations, the labels generated by GPT-4o-mini capture a substantially richer set of fine-grained and previously unannotated objects, including frequently occurring everyday items such as cable, mouse, water bottle, and light switch, as well as many other categories.
Although only Scenes 0–99 are used for training and Scenes 601–610 and 696–705 for testing—amounting to 17.0\% of the total data in \textit{ScanNet-SG-509}—this process yields 3,195 unique object labels, which is 6.3× more than in \textit{ScanNet-SG-509}.
We refer to this benchmark as \textit{ScanNet-SG-GPT}.

\subsection{Subscan-to-subscan (S2S) Matching} \label{Section: S2S Matching}
Each subscan is constructed from 50 to 300 randomly selected yet temporally contiguous frames processed in Section~\ref{section: F2S} (150 to 900 frames in the original ScanNet dataset). We use the method described in Section~\ref{Section: Scene Graph Building} to build the point-cloud map and the scene graph for each subscan. If two subscans from different scans of the same scene overlap, we include the pair as one dataset sample. In total, the dataset contains 13.4k training samples from Scenes 0–599 and 3.4k test samples from Scenes~600–705. In Figure~\ref{fig: cover}B, Subscan 264\_1029 is constructed from frames 264 to 1029 in \texttt{Scene0000\_00} of ScanNet, while Subscan 459\_1027 is constructed from frames 459 to 1027 in \texttt{Scene0000\_01}.

The samples in the S2S task have the following \textbf{difference} compared to the F2S task: 
(i) In F2S, the node-overlap ratio is typically close to one ($>90\%$ on average) from $\mathcal{G}_A$ to $\mathcal{G}_B$ while very low ($< 20\%$ on average) from $\mathcal{G}_B$ to $\mathcal{G}_A$, since the frame-level nodes are usually a subset of those in the complete scan, except when new regions are observed in the cross-scan setting. 
In S2S, the overlap ratio between two subscans spans $(0,1]$. We report the detailed distribution in Fig.~\ref{fig:subscan overlap ratio} in Appendix~\ref{App: OpenSGA Dataset}.
(ii) In F2S, the two scene graphs are expressed in the camera frame and the world frame, respectively, and therefore differ substantially in their coordinate systems. In contrast, in S2S, both scene graphs are expressed in world frames. While these world frames are not identical because the graphs are built from different scans, they share a similar gravity alignment (i.e., the $z$-axis points upward).
(iii) In ScanNet-SG-509, F2S matches a small graph with 4.5 nodes on average to a larger graph with 22.2 nodes on average, whereas S2S matches two graphs with 17.3 nodes on average.
(iv) There are no many-to-one matching in S2S, as both subscans are fused maps.

The general data statistics are presented in Table~\ref{Table: General Statistics1}, where the many-to-one ratio indicates the percentage of nodes that have many-to-one matching from a frame to a scan.
More statistics and visualizations of the dataset are provided in Appendix~\ref{App: OpenSGA Dataset}.

\begin{table}[t]
\centering
\footnotesize 
\caption{Subset Statistics of ScanNet-SG dataset}
\label{Table: General Statistics1}
\setlength{\tabcolsep}{4pt}
\renewcommand{\arraystretch}{0.5}
\begin{tabular}{cllccc}
\toprule
\textbf{Task} &
\multicolumn{1}{c}{\textbf{\begin{tabular}[c]{@{}c@{}}Subset\\Name\end{tabular}}} &
\multicolumn{1}{c}{\textbf{\begin{tabular}[c]{@{}c@{}}Label\\Source\end{tabular}}} &
\textbf{\begin{tabular}[c]{@{}c@{}}Training\\Samples\end{tabular}} &
\textbf{\begin{tabular}[c]{@{}c@{}}Test\\Samples\end{tabular}} &
\textbf{\begin{tabular}[c]{@{}c@{}}M2O\\Ratio\end{tabular}} \\
\midrule
\multirow{2}{*}{\textbf{F2S}} &
\textit{ScanNet-SG-509} & ScanNet & 557.0k & 118.6k & 3.2\% \\
& \textit{ScanNet-SG-GPT} & GPT-4o & 87.9k & 26.9k & 5.0\% \\
\addlinespace
\textbf{S2S} &
\textit{ScanNet-SG-Subscan} & ScanNet & 13.4k & 3.4k & 0.0\% \\
\bottomrule
\end{tabular}
\end{table}

\section{Results} \label{Section: Results}
This section reports experimental results. Section~\ref{section: experiment setup} describes the experimental setup and baseline methods.
In Section~\ref{section: F2S results}, we present results on the F2S alignment task.
Section~\ref{section: F2S multiple scene results} further extends F2S by registering a frame-level SG against multiple scan-level SGs from different scenes, 
thereby simulating matching in large-scale environments containing multiple scenes.
In Section~\ref{section: S2S results}, the results on the S2S task are reported.

Additional results, including additional ablation studies with different input feature types, performance of fine tuned models for the S2S task, performance of models trained jointly on F2S and S2S data, and downstream point cloud registration results, are provided in Appendix~\ref{App: additional results} and Appendix~\ref{App: downstream Relocalization} to maintain clarity and conciseness in the main text.

\subsection{Experimental Setup} \label{section: experiment setup}
Five baseline methods and seven ablation variants of our method are evaluated for F2S and S2S tasks.
As a learning-based baseline, we include SG-Reg (\cite{11024207}), a state-of-the-art scene graph alignment method that leverages instance point clouds, 3D bounding boxes, and text features through a dedicated matching network. 
We further consider four zero-shot baselines. 

The first is ROMAN (\cite{peterson2025roman}), which performs geometrically consistent matching using point cloud features and VLM embeddings. For a fair comparison, we adapt ROMAN by replacing its original FastSAM (\cite{zhao2023fast}) segments with our object-level segments and by using the VLM embeddings generated by GroundingDINO (\cite{liu2024grounding}) rather than CLIP (\cite{radford2021learning}).

The remaining three zero-shot baselines rely on cosine similarity between object embeddings: VLM embeddings from GroundingDINO (\cite{liu2024grounding}), language embeddings from SBERT (\cite{reimers2019sentence}), and their concatenation. For each query object, the instance with the highest similarity exceeding a predefined threshold is selected as the match. These VLM and language embeddings are trained to produce high cosine similarity for semantically consistent objects while separating distinct ones, making cosine similarity a strong yet simple baseline for zero-shot matching.

Finally, we evaluate seven ablation variants of our approach that isolate the impact of different encoders, matchers, and allocators. The full configurations are summarized in Table~\ref{tab:ablation_variants}. 
Our approach builds on the same pretrained features as the zero-shot baselines but further trains a matching network. In general, the model size and computational cost of the variants are primarily determined by the choice of matcher. Variants based on the cosine-similarity efficient matcher are lightweight but typically achieve lower accuracy, whereas variants based on LightGlue3D are more computationally demanding yet consistently deliver stronger performance.
To reflect this trade-off, we report both a \textbf{L}ightweight and a \textbf{H}igh-performance version of our method, and denote them with markers L and H, respectively. In addition to the T-GNN encoder (\cite{11024207}) and our DGSA encoder, we also tested FFN as a baseline variant.

\begin{table}[t]
\caption{Ablation Variants}
\label{tab:ablation_variants}
\centering
\footnotesize 
\setlength{\tabcolsep}{3pt} 
\renewcommand{\arraystretch}{1.1}
\begin{tabular}{lccc}
\toprule
\textbf{Method} & \textbf{Encoder} & \textbf{Matcher} & \textbf{Allocator} \\
\midrule
Ours L T-GNN  + MNN & T-GNN        & Eff. Matcher (L)    & MNN \\
Ours L DGSA + MNN     & DGSA & Eff. Matcher (L)    & MNN \\
Ours L DGSA + MCF \textit{Final}    & DGSA & Eff. Matcher (L)    & MCF      \\
\addlinespace
Ours H FFN + MNN        & FFN               & LightGlue3D (H)     & MNN \\
Ours H T-GNN  + MNN & T-GNN       & LightGlue3D (H)     & MNN \\
Ours H DGSA + MNN     & DGSA & LightGlue3D (H)     & MNN \\
Ours H DGSA + MCF \textit{Final}    & DGSA & LightGlue3D (H)     & MCF      \\
\bottomrule
\end{tabular}
\end{table}

All models are trained on NVIDIA A40 GPUs 
and tested on a desktop equipped with an NVIDIA RTX 3080 Ti. Each model is trained for 10 epochs using the AdamW optimizer (\cite{loshchilov2019decoupled}) with a learning rate of $2\times10^{-4}$ and a batch size of 64. Detailed model parameters are provided in Appendix~\ref{App: Model Parameters}.
Performance is evaluated using four metrics: \textbf{Accuracy}, \textbf{Precision}, \textbf{Recall}, and \textbf{F1 score}. Each metric is computed per test sample over the nodes and then averaged across the test set. The minimum matching score thresholds and other key parameters of the compared methods are optimized using Optuna (\cite{optuna_2019}) to maximize the F1 score.

\subsection{Frame-to-scan Node Matching} \label{section: F2S results}
For this task, all models are trained on the \textit{ScanNet-SG-509} training split, which includes a mixture of self-scan and cross-scan samples. We evaluate performance on three test sets: (i) the self-scan group, (ii) the cross-scan group, and (iii) the \textit{ScanNet-SG-GPT} subset. 
These test sets represent progressively more challenging settings. In (i), query frames are drawn from the same scan used to construct the map, resulting in minimal viewpoint and coverage variation. In (ii), query frames originate from different scanning trajectories, leading to larger viewpoint changes. In (iii), we further evaluate on \textit{ScanNet-SG-GPT}, whose expanded object vocabulary generated by GPT-4o presents the most challenging generalization setting. Importantly, we do not use the \textit{ScanNet-SG-GPT} training split; instead, all models are trained exclusively on \textit{ScanNet-SG-509}. This design enables a direct assessment of cross-vocabulary generalization from \textit{ScanNet-SG-509} to \textit{ScanNet-SG-GPT}

Tables~\ref{tab: results simple}–\ref{tab: results difficult} summarize the F2S matching results in the three test sets. For each metric, the best result is highlighted in bold with an orange background, and the second-best result is highlighted with a cyan background. In general, our method consistently outperforms both SG-Reg (\cite{11024207}) and the zero-shot cosine-similarity baselines. In particular, \textit{Ours H DGSA + MCF Final} achieves the highest accuracy and F1 score in all three settings.

\begin{figure*}
    \centering
    \includegraphics[width=1.0\linewidth]{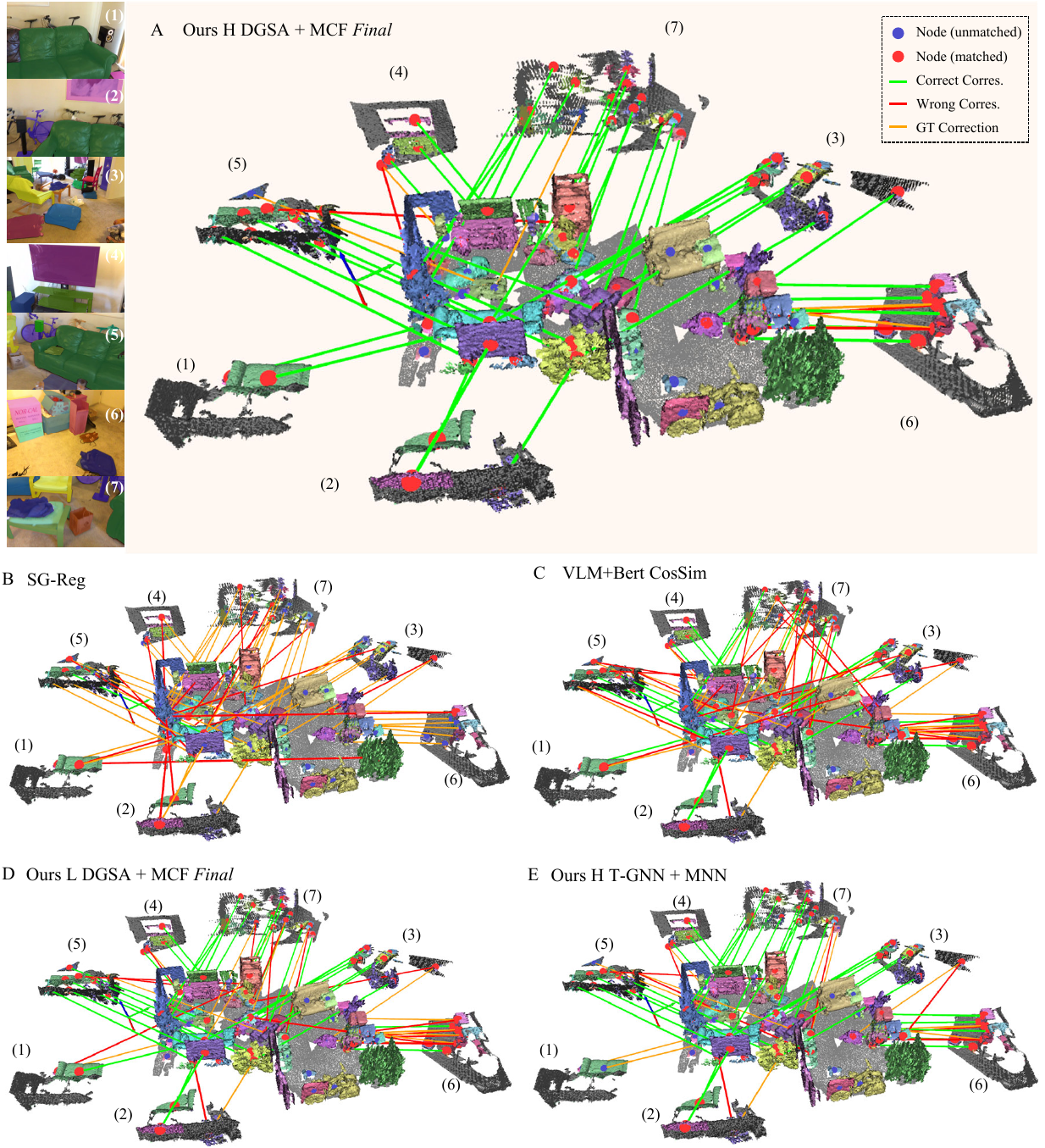}
    \caption{Visualization of matching results produced by different models (A–E). In each subplot, the central point cloud corresponds to the full scan of \texttt{Scene0673\_00}. The query frames are selected from a separate scan of the same scene, \texttt{Scene0673\_01} (cross-scan), and are shown in the top-left corner as instance-segmented RGB images labeled (1)–(7), corresponding to ScanNet frame IDs \texttt{15, 105, 252, 639, 1050, 1521,} and \texttt{1830} (\cite{dai2017scannet}). Objects are visualized with distinct colors, and each object center is marked with a blue dot (unmatched) or a red dot (matched). Correct correspondences are indicated by green lines, while incorrect correspondences are shown in red. Orange lines denote the ground-truth corrected correspondences for either incorrect or missed matches. For visualization purposes, each frame point cloud (originally in its camera coordinate system) is transformed into the world coordinate system of \texttt{Scene0673\_00} and then translated to a nearby location around the scan point cloud. This transformation is applied only for visualization and is not provided to any model. In (A), \textit{Ours H DGSA + MCF} predicts the correct correspondences even when only two and three nodes (corresponding to one and two neighbors) are present in (1) and (2), respectively.}
    \label{fig: F2S reg}
\end{figure*}

\begin{table}[t]
\caption{Matching Results in ScanNet-SG-509 Self Scan Group (Simple)}
\label{tab: results simple}
\centering
\footnotesize 
\setlength{\tabcolsep}{7.5pt}
\renewcommand{\arraystretch}{1}
\begin{tabular}{lcccc}
\toprule
\textbf{Method} & \textbf{Acc $\uparrow$} & \textbf{Pr $\uparrow$} & \textbf{Re $\uparrow$} & \textbf{F1 $\uparrow$} \\
\midrule
SG-Reg            & 0.114 & 0.158 & 0.190 & 0.172 \\
ROMAN           & 0.433 & 0.590 & 0.484 & 0.519 \\
VLM CosSim      & 0.731 & 0.754 & 0.939 & 0.811 \\
Bert CosSim & 0.512 & 0.515 & 0.877 & 0.616 \\
VLM+Bert CosSim & 0.762 & 0.766 & 0.973 & 0.836 \\
\midrule
Ours L T-GNN + MNN & 0.589 & 0.638 & 0.713 & 0.658 \\
Ours L DGSA + MNN              & 0.820 & 0.864 & 0.914 & 0.876 \\

Ours L DGSA + MCF \textit{Final} &
0.838 & 0.838 & \cellcolor{best}\textbf{0.990} & 0.892 \\

Ours H FFN + MNN &
0.877 & 0.890 & 0.958 & 0.912 \\

Ours H T-GNN + MNN  & 0.748 & 0.753 & 0.797 & 0.769 \\

Ours H DGSA + MNN &
\cellcolor{second}0.897 & \cellcolor{best}\textbf{0.906} & 0.967 & \cellcolor{second}0.927 \\

Ours H DGSA + MCF \textit{Final} &
\cellcolor{best}\textbf{0.899} & \cellcolor{second}0.904 & \cellcolor{second}0.974 & \cellcolor{best}\textbf{0.929} \\
\bottomrule
\end{tabular}
\end{table}

\begin{table}[t]
\caption{Matching Results in ScanNet-SG-509 Cross Scan Group (Medium)}
\label{tab: results medium}
\centering
\footnotesize 
\setlength{\tabcolsep}{7.5pt}
\renewcommand{\arraystretch}{1}
\begin{tabular}{lcccc}
\toprule
\textbf{Method} & \textbf{Acc $\uparrow$} & \textbf{Pr $\uparrow$} & \textbf{Re $\uparrow$} & \textbf{F1 $\uparrow$} \\
\midrule
SG-Reg                & 0.027 & 0.039 & 0.042 & 0.040 \\
ROMAN             & 0.358 & 0.414 & 0.374 & 0.380 \\
VLM CosSim        & 0.540 & 0.550 & 0.800 & 0.620 \\
Bert CosSim & 0.455 & 0.458 & 0.715 & 0.519 \\
VLM+Bert CosSim & 0.564 & 0.562 & 0.876 & 0.656 \\
\midrule
Ours L T-GNN  + MNN & 0.481 & 0.499 & 0.667 & 0.548 \\
Ours L DGSA + MNN              & 0.628 & 0.653 & 0.833 & 0.709 \\

Ours L DGSA + MCF \textit{Final} &
0.626 & 0.626 & \cellcolor{best}\textbf{0.925} & 0.723 \\

Ours H FFN + MNN &
0.683 & 0.683 & 0.860 & 0.740 \\

Ours H T-GNN + MNN  & 0.609 & 0.585 & 0.715 & 0.627 \\

Ours H DGSA + MNN &
\cellcolor{second}0.708 & \cellcolor{best}\textbf{0.711} & 0.873 & \cellcolor{second}0.762 \\

Ours H DGSA + MCF \textit{Final} &
\cellcolor{best}\textbf{0.714} & \cellcolor{second}0.709 & \cellcolor{second}0.887 & \cellcolor{best}\textbf{0.768} \\
\bottomrule
\end{tabular}
\end{table}

\begin{table}[t]
\caption{Matching Results in ScanNet-SG-GPT Test Set (Difficult)}
\label{tab: results difficult}
\centering
\footnotesize 
\setlength{\tabcolsep}{7.5pt}
\renewcommand{\arraystretch}{1}
\begin{tabular}{lcccc}
\toprule
\textbf{Method} & \textbf{Acc $\uparrow$} & \textbf{Pr $\uparrow$} & \textbf{Re $\uparrow$} & \textbf{F1 $\uparrow$} \\
\midrule
SG-Reg                & 0.012 & 0.019 & 0.033 & 0.024 \\
ROMAN             & 0.143 & 0.187 & 0.156 & 0.160 \\
VLM CosSim        & 0.306 & 0.354 & 0.504 & 0.381 \\
Bert CosSim  & 0.260 & 0.339 & 0.377 & 0.306 \\
VLM+Bert CosSim & 0.356 & 0.363 & \cellcolor{second}0.740 & 0.462 \\
\midrule
Ours L T-GNN  + MNN & 0.242 & 0.254 & 0.529 & 0.232 \\

Ours L DGSA + MNN &
0.355 & 0.372 & 0.676 & 0.422 \\

Ours L DGSA + MCF \textit{Final} &
0.356 & 0.356 & \cellcolor{best}\textbf{0.788} & 0.466 \\

Ours H FFN + MNN &
0.394 & 0.421 & 0.628 & 0.476 \\

Ours H T-GNN + MNN  & 0.391 & 0.412 & 0.587 & 0.458 \\

Ours H DGSA + MNN &
\cellcolor{second}0.416 & \cellcolor{best}\textbf{0.440} & 0.661 & \cellcolor{second}0.500 \\

Ours H DGSA + MCF \textit{Final} &
\cellcolor{best}\textbf{0.420} & \cellcolor{best}\textbf{0.440} & 0.671 & \cellcolor{best}\textbf{0.503} \\
\bottomrule
\end{tabular}
\end{table}

The first key observation is the importance of pretrained semantic features for open-vocabulary instance matching. The cosine-similarity baselines demonstrate that VLM and SBERT embeddings already provide strong correspondence signals, and their combination further improves performance, indicating that the two modalities capture complementary information.
SG-Reg does not leverage VLM features and instead relies heavily on instance point clouds and coordinate-dependent geometric encoding. 
In F2S, the large coordinate-system mismatch between the frame and the scan, together with partial observations caused by limited view coverage, makes such point-cloud-based features substantially less reliable, leading to a significant performance degradation.
Building on these pretrained representations, our learned models improve robustness across all settings. Even the lightweight variants outperform most baselines, indicating that training enables the matcher to exploit additional contextual and geometric cues beyond raw semantic similarity. We visualize the results of five different methods on seven F2S samples with varying numbers of nodes in the frame in Fig.~\ref{fig: F2S reg}.

A second consistent trend is the impact of the encoder design. T-GNN, originally proposed in SG-Reg (\cite{11024207}), performs noticeably worse than our DGSA encoder across all evaluation settings. 
This is mainly because T-GNN relies on encoding coordinate-sensitive position relations, while F2S involves a significant coordinate-system discrepancy between frame-level and scan-level observations. 
As a result, the geometric patterns learned by T-GNN become difficult to transfer reliably, even compared to the FFN baseline. In contrast, the spatial-attention encoder provides a more stable representation under these conditions, enabling stronger matching performance for both lightweight (L) and high-performance (H) matchers.

\begin{figure*}
     \centering
    \includegraphics[width=1\linewidth]{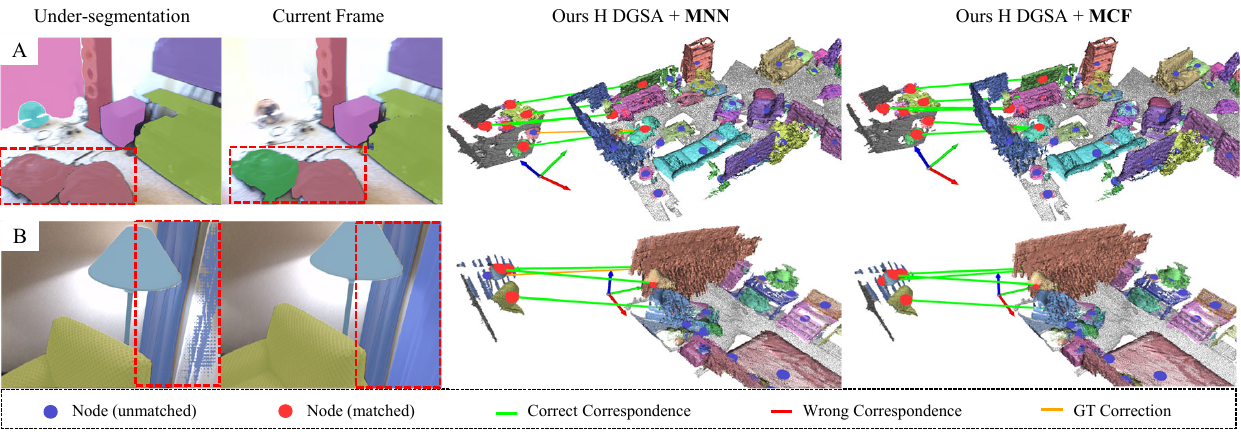}
    \caption{Many-to-one matching illustration. Rows A and C show the instance segmentation and F2S matching results for test samples from \textit{Scene0673}, and \textit{Scene0679}, respectively. The instance segmentation images in the left column illustrate under-segmentation in two representative cases: (A) two adjacent sofas, and (B) two curtain segments. In each case, the highlighted instances (red rectangles) are spatially close and can be segmented either as a single instance or as two separate instances depending on the viewpoint. During scan fusion, the final map often merges these observations into a single instance due to the high geometric overlap and semantic similarity across views. However, when the query frame contains two instances (second column), the matching must correctly resolve a many-to-one correspondence.}
    \label{fig: many-to-one}
\end{figure*}

Comparing our variants with MNN and MCF allocators, we observe that MCF—by explicitly supporting many-to-one correspondences—primarily improves performance by increasing recall and yielding more stable F1 scores across all test sets. With the lightweight matcher, recall and F1 improve by 0.093 and 0.025 on average over the three test sets (from Ours L DGSA + MNN to Ours L DGSA + MCF Final). With the high-performance matcher, the gains are smaller, with average improvements of 0.010 in recall and 0.003 in F1 (from Ours H DGSA + MNN to Ours H DGSA + MCF Final).
The larger improvement in the lightweight setting stems from the fact that the lightweight matcher relies mainly on feature similarity and does not enforce geometric consistency between matched object pairs, whereas MCF incorporates such global allocation constraints. In contrast, the high-performance matcher already models relative-distance consistency internally, and therefore, MCF mainly contributes by enabling many-to-one matching. As shown in Table~\ref{Table: General Statistics1}, many-to-one cases occur in only a small fraction of nodes in the dataset, which limits the overall performance gain. In Figure~\ref{fig: many-to-one}, we compare the matching results using the MNN and MCF allocators for samples with many-to-one correspondences. MNN fails to resolve the many-to-one case, causing one of the two instances to be either mismatched or assigned to the no-match class. In contrast, MCF correctly handles the many-to-one correspondence and produces consistent matches.

\begin{figure}
    \centering
    \includegraphics[width=1\linewidth]{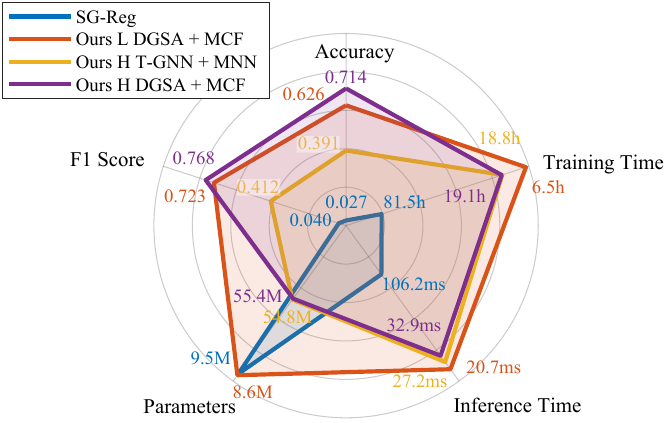}
    \caption{Radar plot comparing four learning-based methods across five metrics on F2S task. Accuracy and F1 score are shown on a $[0,1]$ scale. Parameters, training time (per epoch on an NVIDIA A40 GPU), and inference time (per sample on an NVIDIA 3080Ti GPU) are normalized to the ranges $[5,100]$~M, $[5,100]$~h, and $[10,150]$~ms, respectively.}
    \label{fig: radar plot F2s}
\end{figure}

To jointly compare efficiency and matching performance, we visualize the results in a radar plot in Fig.~\ref{fig: radar plot F2s}. Efficiency is quantified by the number of model parameters, the average training time per epoch, and the average inference time per sample, while performance is measured by accuracy and F1 score. All results are reported on the cross-scan (\emph{Medium}) test set. 
The plot includes SG-Reg and three representative variants of our method. We omit the cosine-similarity baselines, as they require no additional training and their inference cost is dominated by a simple similarity computation.
As shown in Fig.~\ref{fig: radar plot F2s}, our variants are three times faster than SG-Reg. \textit{Ours L DGSA + MCF Final} achieves a favorable trade-off between efficiency and performance, whereas \textit{Ours H DGSA + MCF Final} is larger and slower but has the best overall accuracy and F1 score.

\subsection{Frame-to-scan with Multiple Scenes} \label{section: F2S multiple scene results}

When operating in a large-scale environment containing multiple scenes, the system must identify the correct scene against which to perform object node matching.
To showcase the necessity of a global embedding, we evaluate the retrieval accuracy of the correct scene graph from a graph database given a query graph obtained from an image. 
The metric Recall@K is employed to evaluate whether the target scene is ranked among the top K scenes in terms of similarity. 
Additionally, we report the inference time to assess the efficiency of each method.
For this evaluation, we use the data from the testing scenes (scenes 600-705) and treat each scene's graph as one entry of our database, ensuring that there is only one correct graph in our target set. In these experiments, we only evaluate the final versions of our networks: the high-performance variant (Ours H) and the lightweight variant (Ours L), both employing DGSA and MCF.

The first experiment evaluates the ability of the proposed global embedding to retrieve the correct graph from a database of graphs. We test different values of K to evaluate the capacity of using the global embedding as a filtering step in the retrieval step. 
The results of this experiment are presented in Table \ref{tab:retrieval@k}. As it can be seen, both variants yield similar retrieval accuracy across different K value, mainly because they employ the same number of multi-head attention layers to produce the global embedding.

\begin{table}[t]
\caption{Retrieval accuracy with different K values when using the global scene descriptors.}
\label{tab:retrieval@k}
\centering
\footnotesize 
\setlength{\tabcolsep}{5pt}
\renewcommand{\arraystretch}{1.15}
\begin{tabular}{lcccc}
\toprule
\textbf{Method} & \textbf{Recall@1} $\uparrow$ & \textbf{Recall@5} $\uparrow$ & \textbf{Recall@10} $\uparrow$ & \textbf{Recall@20} $\uparrow$\\
\midrule
Ours L  & 0.63 & 0.89 & 0.95 & 0.98 \\
Ours H  & 0.63 & 0.88 & 0.95 & 0.98 \\
\bottomrule
\end{tabular}
\end{table}

For our final experiment, we ablate different methods to obtain the correct scene graph (Recall@1) from one forward pass with our network: match against all candidates followed by reranking (\textit{baseline}), or first using the global embedding for filtering, denoted as top K, and then performing reranking. In our experiments, we also compare two reranking strategies: the weighted reranking proposed in equation (\ref{eq:reranking}) (\textit{weight rerank}) and direct reranking based on the matching score, which would be the same as in (\ref{eq:reranking}) but with the dot product being equal to one.
The filtering step, based on the global embedding, consists of selecting the Top-K candidates from the database according to the cosine similarity of the global embeddings, after which only the selected scenes are reranked. The results are presented in Table~\ref{tab:match_retrieval}.

From the results, we observe that the high-performance version of our network achieves higher accuracy at the expense of increased computation time. Compared to the baseline, incorporating filtering improves overall accuracy by removing incorrect graphs that would otherwise obtain high matching scores.
Regarding the reranking strategies, the proposed weighted reranking consistently outperforms score-based reranking, indicating that the global embeddings of correct matches are closer in the embedding space than those of incorrect pairs.
In terms of processing time, all methods that use the global embedding for filtering are more efficient than performing individual graph matching for all candidates, followed by a selection. As the filter size (K) increases, computation time increases, while accuracy also improves. As expected, the lightweight version is more computationally efficient than the high-performance version of the network.

\begin{table}[t]
\caption{Correct scene graph retrieval and computation time in ScanNet-SG Test Set (Medium) with different reranking strategies.}
\label{tab:match_retrieval}
\centering
\footnotesize 
\setlength{\tabcolsep}{11pt}
\renewcommand{\arraystretch}{1}
\begin{tabular}{lcc}
\toprule
\textbf{Method} & \textbf{Recall@1 $\uparrow$} & \textbf{Time $\downarrow$} \\
\midrule
Ours L + \textit{baseline} & 0.64 & 0.057 \\
Ours L + Top 5 + \textit{direct rerank} & 0.60 & 0.004 \\
Ours L + Top 10 + \textit{direct rerank} & 0.64 & 0.011 \\
Ours L + Top 20 + \textit{direct rerank} & \cellcolor{second}0.68 & 0.015 \\
Ours L + Top 5 + \textit{weighted rerank} & 0.62 & 0.005 \\
Ours L + Top 10 + \textit{weighted rerank} & 0.66 & 0.011 \\
Ours L + Top 20 + \textit{weighted rerank} & \cellcolor{best}\textbf{0.69} & 0.015 \\
\midrule
Ours H + \textit{baseline} & 0.58 & 0.835 \\
Ours H + Top 5 + \textit{direct rerank} & 0.67 & 0.048 \\
Ours H + Top 10 + \textit{direct rerank} & 0.67 & 0.091 \\
Ours H + Top 20 + \textit{direct rerank} & 0.66 & 0.181 \\
Ours H + Top 5 + \textit{weighted rerank} & 0.71 & 0.048 \\
Ours H + Top 10 + \textit{weighted rerank} & \cellcolor{second}0.73 & 0.091 \\
Ours H + Top 20 + \textit{weighted rerank} & \cellcolor{best}\textbf{0.74} & 0.181 \\
\bottomrule
\end{tabular}
\end{table}

\subsection{Subscan-to-subscan Node Matching} \label{section: S2S results}
\begin{figure*}
    \centering
    \includegraphics[width=1\linewidth]{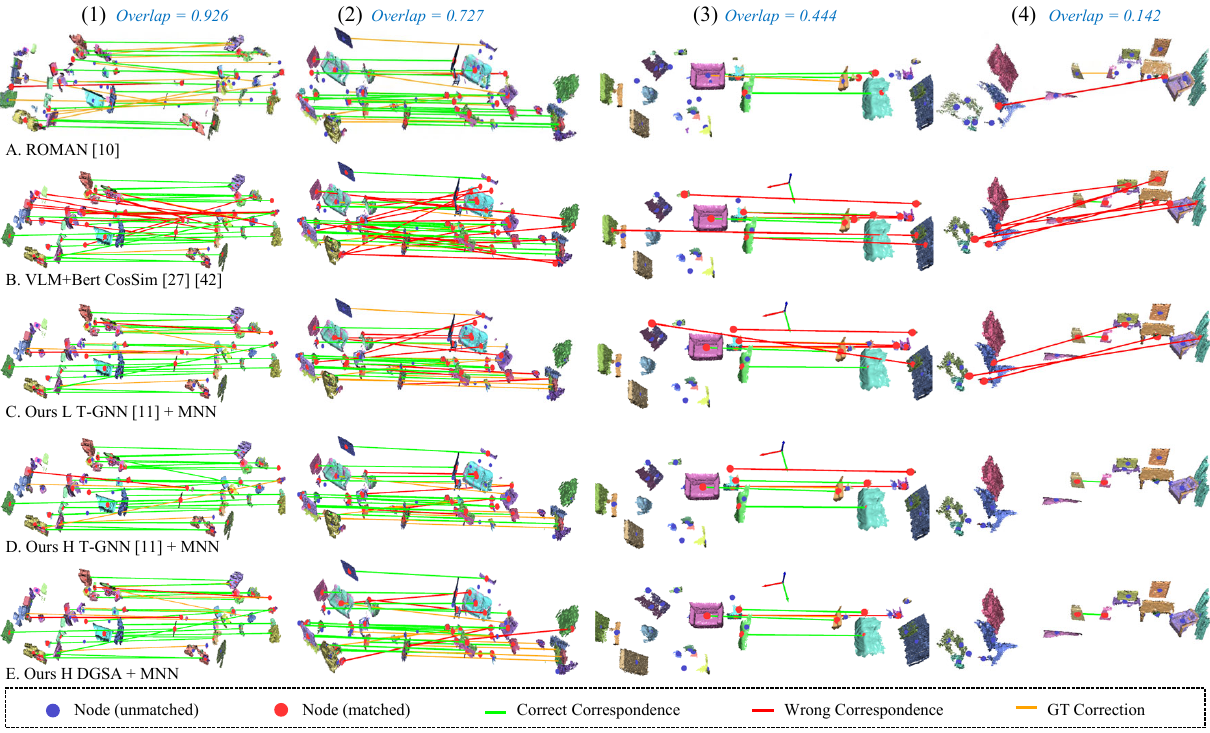}
    \caption{Comparison of different methods (A--E) on the S2S alignment task. The subscan pairs in (1)–-(4) are: (1) \texttt{Scene0673\_00} frames 1383–2139 and \texttt{Scene0673\_01} frames 1584–1989; (2) \texttt{Scene0673\_00} frames 654–1140 and \texttt{Scene0673\_02} frames 234–1032; (3) \texttt{Scene0698\_00} frames 732–942 and \texttt{Scene0698\_01} frames 837–1302; and (4) \texttt{Scene0642\_00} frames 168–777 and \texttt{Scene0642\_01} frames 1890–2367. The node overlap ratio between two subscans is shown with blue text on the left. The overlap ratio decreases from 0.926 to 0.142 from (1) to (4). Points belonging to different instances are rendered in random colors. Background point clouds are omitted for clarity.}
    \label{fig: s2s results}
\end{figure*}

Table~\ref{tab: s2s results} presents the results of the S2S node matching task. The results of two high-performing baseline methods and three of our variants are further illustrated in Fig.~\ref{fig: s2s results}.
Since both subscans are fused from multiple frames, the S2S setting contains almost no many-to-one correspondences (Table~\ref{Table: General Statistics1}). Therefore, we use MNN as the only allocator and focus on comparing different feature choices and matching network designs. 

In the S2S node matching task, the zero-shot cosine-similarity baselines achieve relatively high recall, especially for VLM CosSim (0.906 Re) and VLM+Bert CosSim (0.898 Re). However, their precision remains low (0.370–0.400), leading to moderate F1 scores (0.477–0.530). 
In contrast, our learned matching models substantially improve accuracy and precision, demonstrating that training a matcher on top of pretrained semantics is critical for disambiguating semantically similar objects. Compared to the best cosine-similarity baseline (VLM+Bert CosSim, 0.530 F1), all variants of our method achieve stronger overall performance. In particular, the best-performing model \textit{Ours H DGSA + MNN} achieves 0.681 Acc and 0.589 F1, reflecting a more balanced precision–recall trade-off.

Unlike in F2S, T-GNN (\cite{11024207}) becomes a more competitive encoder in S2S. 
In this task, both graphs are constructed from fused multi-frame subscans, which reduces the coordinate-system mismatch and alleviates the extreme partial-observation issue present in F2S. 
As a result, T-GNN-based variants benefit more from their geometric encoding and achieve strong performance: \textit{Ours H T-GNN + MNN} reaches the best F1 (0.592) and the highest recall among our variants (0.732). 
Meanwhile, the FFN encoder yields the highest precision (0.565) but lower recall (0.592), leading to a lower F1 (0.556). 
This suggests that in S2S, geometric consistency encoded by T-GNN helps recover additional true correspondences, whereas simpler encoders tend to be more conservative. 
SG-Reg (\cite{11024207}) remains significantly behind all other methods. 
We observe that in the training set, SG-Reg achieves a precision of 0.501 and a recall of 0.512, indicating a tendency to overfit to training data. Although SG-Reg proposed T-GNN, its matching network and feature design rely primarily on point cloud information and do not incorporate VLM features. 
This design is not suited for the subscan matching setting, where instance point clouds are often incomplete and the overlap between subscans is limited.
ROMAN (\cite{peterson2025roman}), which additionally incorporates PCA-based point cloud features alongside VLM embeddings, achieves relatively high precision but substantially lower recall compared to the other zero-shot baselines. As a result, its overall F1 score is the lowest among the four zero-shot methods, primarily due to its limited recall.

\begin{table}[t]
\caption{Subscan-to-subscan Matching Result}
\label{tab: s2s results}
\centering
\footnotesize 
\setlength{\tabcolsep}{7.5pt}
\renewcommand{\arraystretch}{1}
\begin{tabular}{lcccc}
\toprule
\textbf{Method} & \textbf{Acc $\uparrow$} & \textbf{Pr $\uparrow$} & \textbf{Re $\uparrow$} & \textbf{F1 $\uparrow$} \\
\midrule
SG-Reg             & 0.056 & 0.171 & 0.078 & 0.107 \\
ROMAN   & 0.430 & 0.505 & 0.268 & 0.336 \\
VLM CosSim      & 0.461 & 0.370 & \cellcolor{best}\textbf{0.906} & 0.502 \\
Bert CosSim        & 0.535 & 0.382 & 0.751 & 0.477 \\
VLM+Bert CosSim        & 0.518 & 0.400 & \cellcolor{second}0.898 & 0.530 \\
\midrule
Ours L T-GNN + MNN & 0.642 & 0.502 & 0.713 & 0.568 \\
Ours L DGSA + MNN     & 0.628 & 0.490 & 0.696 & 0.554 \\

Ours H FFN + MNN        &
\cellcolor{second}0.677 & \cellcolor{best}\textbf{0.565} & 0.592 & 0.556 \\

Ours H T-GNN + MNN &
0.667 & 0.525 & 0.732 & \cellcolor{best}\textbf{0.592} \\

Ours H DGSA + MNN     &
\cellcolor{best}\textbf{0.681} & \cellcolor{second}0.547 & 0.684 & \cellcolor{second}0.589 \\
\bottomrule
\end{tabular}
\end{table}

\begin{figure}
    \centering
    \includegraphics[width=1\linewidth]{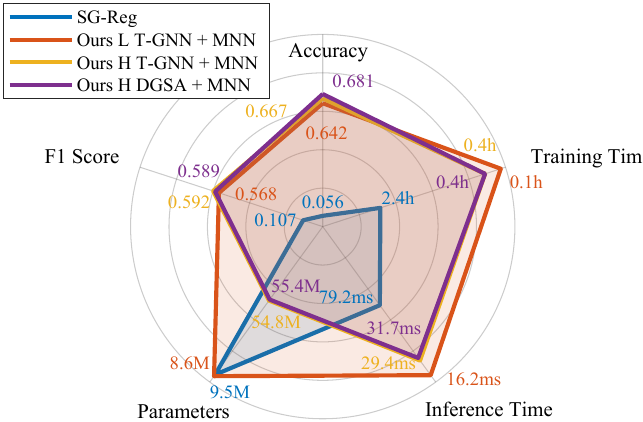}
    \caption{Radar plot comparing four learning-based methods across five metrics on S2S task. Accuracy and F1 score are shown on a $[0,1]$ scale. Parameters, training time (per epoch on an NVIDIA A40 GPU), and inference time (per sample on an NVIDIA 3080Ti GPU) are normalized to the ranges $[5,100]$~M, $[0.01,3.5]$~h, and $[10,150]$~ms, respectively. }
    \label{fig:radar plot s2s}
\end{figure}

\begin{figure}
    \centering
    \includegraphics[width=1\linewidth]{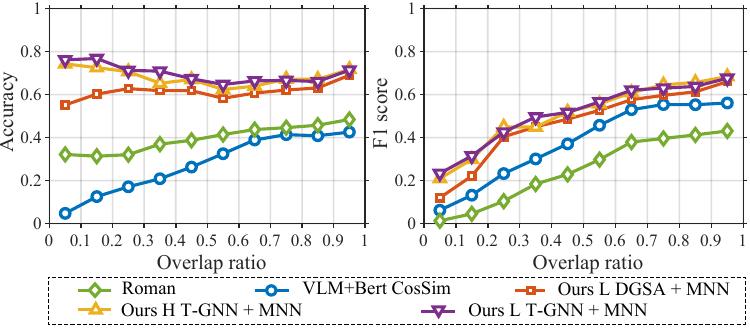}
    \caption{Line plot of accuracy (left) and F1 score (right) changing with different overlap ratio of two subscans.}
    \label{fig:subscan_line_plot}
\end{figure}

Moreover, we find that the performance of all methods strongly depends on the overlap ratio between two subscans. In Figure~\ref{fig:subscan_line_plot}, we further group the test set into 10 bins according to overlap ratio from 0 to 1 with a step size of 0.1. In the F1 score–overlap ratio plot, the F1 score of all methods increases noticeably with overlap, indicating that the amount of shared observed area is a primary factor affecting subscan matching difficulty.

Interestingly, the accuracy curves of our variants remain relatively flat compared to their F1 curves, whereas the zero-shot VLM+Bert CosSim (\cite{liu2024grounding,reimers2019sentence}) baseline shows a much steeper increase in accuracy as overlap grows. This behavior is mainly driven by the handling of no-match cases at low overlap. 
When the overlap is small, most query objects have no correspondence in the other subscan, making true negatives dominate the evaluation. 
While true negatives are reflected in accuracy, they do not contribute to the F1 score.
ROMAN (\cite{peterson2025roman}) and our models, which consider geometric consistency, are more reliable at predicting the “no-match” label in these low-overlap regimes, maintaining stable accuracy even when the overlap is limited. 
In contrast, VLM+Bert CosSim tends to produce spurious matches when no correspondence exists due to its lack of geometric consistency, leading to many false positives and thus lower accuracy at low overlap. 
Notably, compared to the zero-shot setting of ROMAN, the additional task-specific training of our models enables the matcher to better adapt the feature representations and more accurately predict the “no-match” class, thereby improving accuracy under low-overlap conditions.

Overall, the S2S results confirm that subscan-level matching benefits from additional training. Compared to F2S, the T-GNN encoder is more effective in S2S, which can be attributed to the reduced coordinate discrepancy between subscans.
We further summarize the performance and efficiency metrics of the trained methods in the radar plot shown in Fig.~\ref{fig:radar plot s2s}. Considering the trade-off between accuracy and computational cost, \textit{Ours L T-GNN (\cite{11024207}) + MNN} provides the best overall balance, whereas \textit{Ours H DGSA + MNN} achieves the highest accuracy.

Although the S2S task differs from the F2S task as discussed in Section~\ref{Section: S2S Matching}, both tasks involve object node matching using the same types of features. Therefore, pretraining the model on the F2S dataset, which is substantially larger than the S2S dataset, and subsequently fine-tuning it on the S2S dataset is expected to improve performance. We report the results in Appendix~\ref{App: FT test}, which show that fine-tuning improves accuracy by 2–3\% and F1 score by more than 4\% compared to training solely on the S2S dataset.
In addition, Appendix~\ref{App: all data test} presents results for models trained using all available training data, including F2S data annotated with ScanNet labels and GPT-4o labels, as well as S2S data. The results show improved performance on the S2S task and on the difficult F2S test group (GPT-4o annotated ). In contrast, a performance decrease is observed on the easy and medium F2S groups annotated with ScanNet labels since more labels are involved. Detailed results are provided in the appendix.


\section{Conclusion} \label{Section: Conclusion}

This paper presented a learning-based framework and a benchmark for scene graph alignment under partial observations, covering both F2S and S2S node matching. Across both tasks and all three F2S evaluation settings, our methods consistently achieve the best overall performance.
The results highlight two main insights. 
First, features derived from raw instance point clouds are unreliable under partial observations and coordinate discrepancies, especially in F2S, where point-cloud-centric designs degrade substantially.
In contrast, pretrained semantic representations are crucial: VLM and SBERT embeddings already provide strong zero-shot matching signals, and combining them yields complementary gains of 2.5\%–8.1\% in F1 score across evaluation settings.
Second, learning on top of these pretrained features further improves robustness. Our trained matching network consistently outperforms zero-shot cosine-similarity baselines by reducing semantic ambiguity and incorporating contextual and geometric cues, improving accuracy by 6.4\%–13.7\% and F1 score by 4.1\%–11.2\%.
Compared to the training-based baseline, our method further reduces both training and inference time by over 60\%.

Despite these strengths, our current dataset still has limitations. In particular, edge relationships are mainly represented by geometric distance, and their language descriptions are simplified to a single “next to” relation, limiting the evaluation of richer relation-aware grounding.
Enriching edge descriptions and expanding relation diversity are promising directions for future work, alongside further exploiting multi-modal cues to improve alignment in larger-scale and outdoor environments.

\section{Funding}
This work is funded in part by the European Union (ERC, INTERACT, 101041863). Views and opinions expressed are however those of the author(s) only and do not necessarily reflect those of the European Union or the European Research Council Executive Agency. Neither the European Union nor the granting authority can be held responsible for them.

\appendix
\section{Appendix}
\subsection{Details of Multiview Fusion} \label{App: Merging}
A key technique in multiview fusion is instance tracking across input frames. When ScanNet provides semantic labels and coarse instance segmentations projected from the 3D mesh with manually assigned instance IDs, as in the construction of \textit{ScanNet-SG-509}, we associate fine-grained instances obtained from GroundedSAM with ScanNet coarse instances using IoU-based matching to inherit consistent instance IDs. Points belonging to the same instance are then merged and downsampled using a voxel grid filter with a resolution of 0.02 m.

For open-set labels tagged by GPT-4o, where no ground-truth instance IDs are available, we perform instance tracking based on both semantic and geometric consistency. Specifically, two instances are considered identical if their SBERT embedding cosine similarity and their voxel-level 3D IoU both exceed predefined thresholds. SBERT embeddings help distinguish semantically different objects that may have high geometric overlap, such as a tablecloth on a table or a book on a laptop. This matching condition is first applied sequentially across frames to construct an initial instance-aware point cloud map for each scan. We then iteratively compare all existing instances and merge those satisfying the same condition until convergence. To improve efficiency, a center distance check is applied prior to similarity evaluation to filter unlikely candidates. This post-merging step is particularly important for resolving duplicate instances caused by multiple times of scans of the same object.

Although GroundedSAM leverages the foundation model SAM for segmentation, segmentation noise near object boundaries can introduce outlier points in the fused map. To address this issue, we further apply DBSCAN (\cite{ester1996density}) and Statistical Outlier Removal (SOR) filtering, retaining only the largest connected point cluster for each instance to improve the quality of the reconstructed object point clouds.

\subsection{Dataset Statistics and Visualization} \label{App: OpenSGA Dataset}
In Section~\ref{Section: Scene Graph Building and Dataset Construction}, we describe the dataset construction pipeline.
We present the registration performance for cross-scan transformation matrix estimation (described in Section~\ref{section: F2S}) in Tab.~\ref{tab:registration_quality}. Thresholds of 0.05 m and 0.1 m are used to compute the mean and median fitness, RMSE, and overlap ratio (OR), respectively.

\begin{figure*}
    \centering
    \includegraphics[width=1\linewidth]{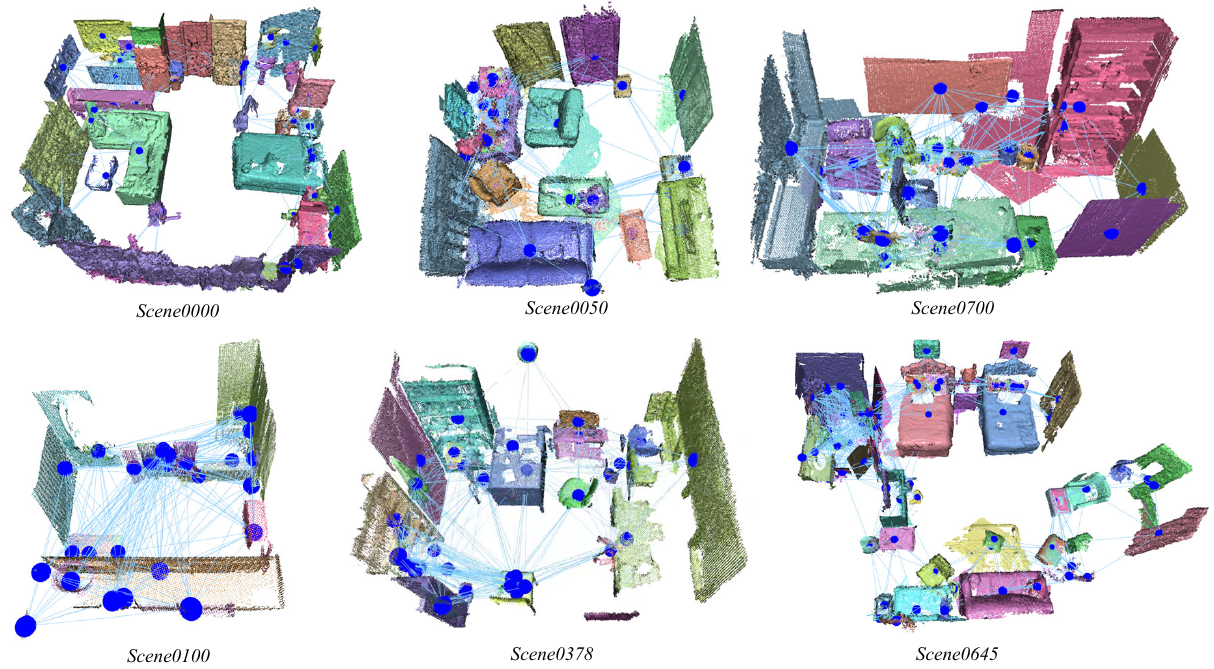}
    \caption{Examples of 3D scene graphs constructed using ScanNet human-annotated object labels (\textit{ScanNet-SG-509}). The blue points show the centers of objects. An edge is added to connect two objects if their distance is smaller than two meters.}
    \label{fig: dataset 509}
\end{figure*}

\begin{figure*}
    \centering
    \includegraphics[width=1\linewidth]{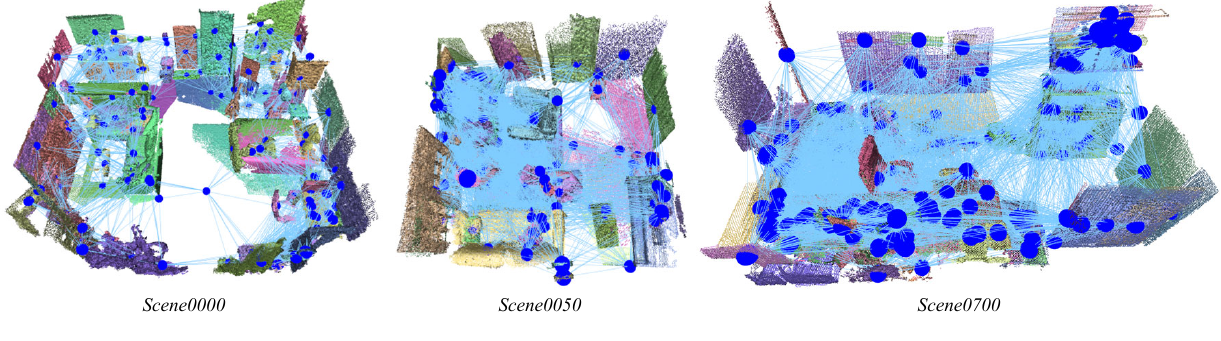}
    \caption{Examples of 3D scene graphs constructed using GPT-4o–tagged objects \textit{ScanNet-SG-GPT}.
The three scenes correspond to those shown in the first row of Fig.~\ref{fig: dataset 509}, but contain substantially more object instances and a significantly larger set of object classes.}
    \label{fig: dataset gpt}
\end{figure*}

Figures~\ref{fig: cover}, \ref{fig: F2S reg}, and \ref{fig: s2s results} present examples of frame-, subscan-, and scan-level 3D scene graphs built using ScanNet labels, together with their corresponding matching results. Figure~\ref{fig: dataset 509} further shows object instances and scene graphs from six scenes of different sizes constructed with ScanNet labels (\textit{ScanNet-SG-509} group). For comparison, Fig.~\ref{fig: dataset gpt} illustrates the corresponding scene graphs for the first three scenes when GPT-4o is used for object tagging (\textit{ScanNet-SG-GPT} group), where substantially more object nodes are observed.

\begin{table}[t]
\caption{Registration Performance for Cross-scan Matching}
\label{tab:registration_quality}
\centering

\scriptsize
\setlength{\tabcolsep}{4pt}
\renewcommand{\arraystretch}{1.2}

\begin{tabular}{lcccccc}
\toprule

\multirow{2}{*}{\textbf{THR}} 
& \textbf{Mean} 
& \textbf{Med.} 
& \textbf{Mean} 
& \textbf{Med.} 
& \textbf{Mean} 
& \textbf{Med.} \\

& \textbf{Fitness$\uparrow$} 
& \textbf{Fitness$\uparrow$} 
& \textbf{RMSE$\downarrow$ (cm)} 
& \textbf{RMSE$\downarrow$ (cm)} 
& \textbf{OR$\uparrow$} 
& \textbf{OR$\uparrow$} \\

\midrule

0.05 m 
& 0.750 
& 0.798 
& 2.29 
& 2.30 
& 0.800 
& 0.840 \\

0.10 m 
& 0.844 
& 0.899 
& 3.21 
& 3.10 
& 0.898 
& 0.938 \\

\bottomrule
\end{tabular}
\end{table}

\begin{figure*}
    \centering
    \includegraphics[width=1\linewidth]{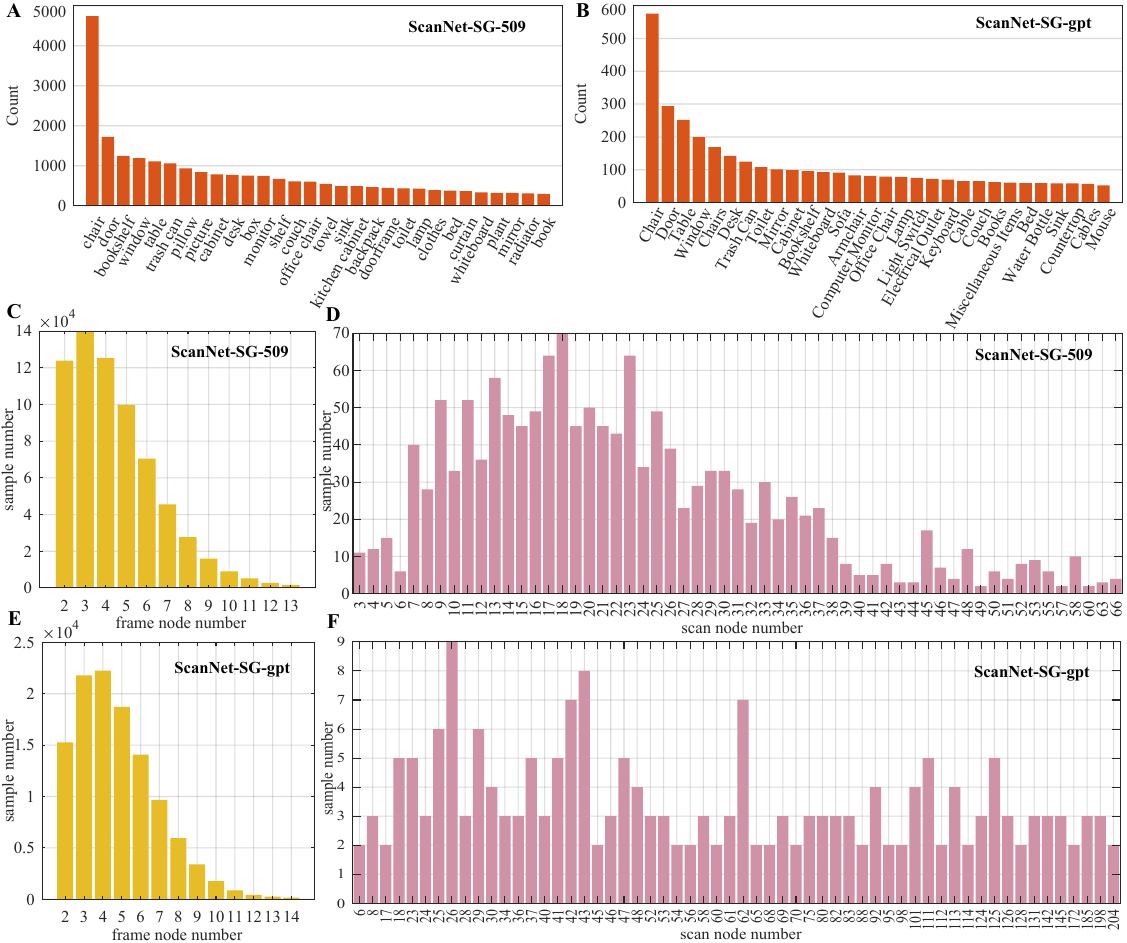}
    \caption{Dataset statistics for the frame-to-scan (F2S) matching task. (A–B) Top-30 most frequent object categories in the \textit{ScanNet-SG-509} and \textit{ScanNet-SG-GPT} groups, respectively. (C–D) Distribution of node counts in frame-level and scan-level scene graphs for the \textit{ScanNet-SG-509} group. (E–F) Corresponding node count distributions for the \textit{ScanNet-SG-GPT} group.}
    \label{fig: node num}
\end{figure*}

\begin{figure*}
    \centering
    \includegraphics[width=1.0\linewidth]{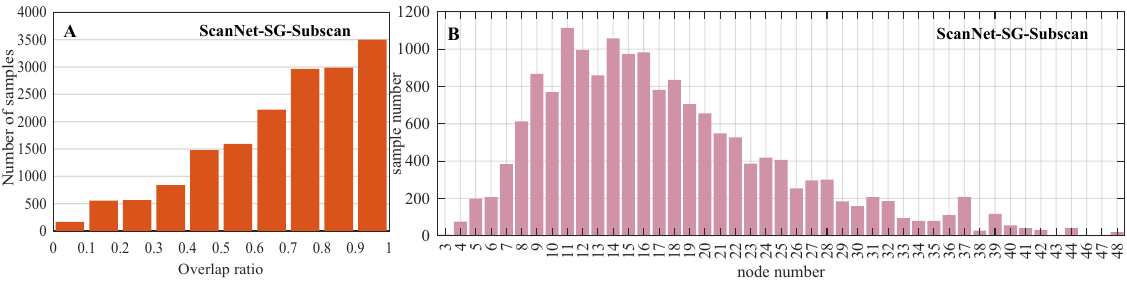}
    \caption{Dataset statistics for the subscan-to-subscan (S2S) matching task (\textit{ScanNet-SG-Subscan group}). (A) Distribution of node overlap ratio of two matched subscans. (B) Distribution of node counts in subscans.}
    \label{fig:subscan overlap ratio}
\end{figure*}

The top 30 most frequent object categories in the \textit{ScanNet-SG-509} and \textit{ScanNet-SG-GPT} groups are shown in Fig.~\ref{fig: node num}(A) and (B), respectively. We further present the distributions of node counts in frame-level and scan-level scene graphs for the \textit{ScanNet-SG-509} and \textit{ScanNet-SG-GPT} groups in Fig.~\ref{fig: node num}(C)–(F). The number of nodes ranges from 3 to 66 in the \textit{ScanNet-SG-509} group and from 6 to 204 in the \textit{ScanNet-SG-GPT} group, reflecting the increased scene complexity enabled by open-set labeling.

Each subscan in the \textit{ScanNet-SG-Subscan} group represents a fragment of a corresponding scan in the \textit{ScanNet-SG-509} group. The node count distribution of subscan scene graphs is shown in Fig.~\ref{fig:subscan overlap ratio}(B), while Fig.~\ref{fig:subscan overlap ratio}(A) presents the overlap ratio distribution for subscan matching pairs.

\subsection{Model Parameters}\label{App: Model Parameters}
This section provides values of the parameters used in our methods. The DGSA encoder uses 8 attention heads, 4 layers, and a dropout rate of 0.1. The fused feature dimension for both LightGlue3D and the efficient matcher is 512.
In the MCF algorithm, the key parameters are $\tau=0.3$, $K=5$, $\text{C}_{unmatched}=2.0$, $\lambda=1.0$, and a maximum of 5 iterations.
We assume that the maximum number of nodes in a frame that can correspond to a single node in the map is unknown; therefore, $\text{Cap}_{max}$ is set to unlimited.
For MNN, the minimum similarity threshold is set to 0.2 for the LightGlue3D (H) version and 0.1 for the efficient matcher (L) version.

\subsection{Additional Results} \label{App: additional results}
This section provides additional scene graph alignment results as complementary evaluations to the main experiments.

\subsubsection{Ablation Study with Different Input Descriptors} \label{App: ablation study}

Three types of features are used in our experiments: the VLM embedding $\boldsymbol{f}{vl}$ extracted from GroundingDINO (\cite{liu2024grounding}), the text embedding $\boldsymbol{f}{t}$ obtained from SBERT (\cite{reimers2019sentence}), and the geometric feature $\boldsymbol{f}_{\mathrm{g}}$ derived from 3D bounding box attributes. As shown in Table~\ref{tab: ablation different features}, we train \textit{Ours L DGSA + MNN} and \textit{Ours H DGSA + MNN} using different feature combinations and evaluate them on the cross-scan (Medium) group of the F2S task.

Using only the VLM embedding $\boldsymbol{f}{vl}$ already achieves strong performance for both models, confirming that VLM features provide highly discriminative representations for object matching. Incorporating SBERT text embeddings $\boldsymbol{f}{t}$ consistently improves performance, yielding higher accuracy and F1 scores for both the lightweight and high-capacity models.
Adding geometric features $\boldsymbol{f}{g}$ also improves performance compared to using $\boldsymbol{f}{vl}$ alone, although the gain is smaller than that from text embeddings. When all three features are used jointly, both models achieve their highest accuracy and precision, with \textit{Ours H DGSA + MNN} also achieving the best overall F1 score of 0.762. However, for \textit{Ours L DGSA + MNN}, incorporating all features slightly reduces recall, suggesting that the lightweight model becomes more conservative when using richer feature representations.

Overall, these results demonstrate that fusing all three feature types provides the most effective solution, improving matching precision and overall robustness by leveraging complementary semantic and geometric information.

\begin{table}[t]
\centering
\footnotesize 
\caption{Matching Result with Different Feature Sources (Medium)}
\label{tab: ablation different features}
\setlength{\tabcolsep}{3.5pt}
\renewcommand{\arraystretch}{1.15}
\begin{tabular}{clcccc}
\toprule
\textbf{Method} & \textbf{Features} & \textbf{Acc $\uparrow$} & \textbf{Pr $\uparrow$} & \textbf{Re $\uparrow$} & \textbf{F1 $\uparrow$} \\
\midrule
\multirow{4}{*}{Ours L DGSA + MNN} & $\boldsymbol{f}_\mathrm{vl}$ & 0.614 & 0.614 & 0.921 & 0.713 \\
& $\boldsymbol{f}_\mathrm{vl} + \boldsymbol{f}_{\mathrm{t}}$ & 0.621 & 0.621 & \cellcolor{best}\textbf{0.923} & \cellcolor{best}\textbf{0.718} \\
& $\boldsymbol{f}_\mathrm{vl} + \boldsymbol{f}_{\mathrm{g}}$ & 0.620 & 0.619 & 0.922 & 0.717 \\
& $\boldsymbol{f}_\mathrm{vl} + \boldsymbol{f}_{\mathrm{t}} + \boldsymbol{f}_{\mathrm{g}}$ & \cellcolor{best}\textbf{0.628} & \cellcolor{best}\textbf{0.653} & 0.833 & 0.709 \\
\midrule
\multirow{4}{*}{Ours H DGSA + MNN} & $\boldsymbol{f}_\mathrm{vl}$ & 0.690 & 0.691 & \cellcolor{best}\textbf{0.875} & 0.751 \\
& $\boldsymbol{f}_\mathrm{vl} + \boldsymbol{f}_{\mathrm{t}}$ & 0.706 & 0.706 & 0.871 & 0.758 \\
& $\boldsymbol{f}_\mathrm{vl} + \boldsymbol{f}_{\mathrm{g}}$ & 0.695 & 0.697 & 0.873 & 0.754 \\
& $\boldsymbol{f}_\mathrm{vl} + \boldsymbol{f}_{\mathrm{t}} + \boldsymbol{f}_{\mathrm{g}}$ & \cellcolor{best}\textbf{0.708} & \cellcolor{best}\textbf{0.711} & 0.873 & \cellcolor{best}\textbf{0.762} \\
\bottomrule
\end{tabular}
\end{table}

\subsubsection{S2S Results with Finetuned Model From F2S}  \label{App: FT test}
We fine-tune our \textit{Ours L DGSA + MNN} and \textit{Ours H DGSA + MNN} models, which were pre-trained on the F2S dataset as described in Section~\ref{section: F2S results}, using the training data from the S2S task and evaluate them on the corresponding S2S test set. The results are presented in Table~\ref{tab:s2s_ft_results}. The gain is computed relative to the corresponding results reported in Table~\ref{tab: s2s results}.

The results show that fine-tuning consistently improves performance across all evaluation metrics for both models. In particular, all metrics improve by at least 2.4\%, with recall showing the most significant gain, increasing by more than 5\% for both models. This demonstrates that pretraining on the larger F2S dataset provides strong initialization, while task-specific fine-tuning further adapts the model to the characteristics of the S2S task, resulting in improved matching performance.

\begin{table}[t]
\caption{Subscan-to-subscan Matching Result with Fine-tuned Models}
\label{tab:s2s_ft_results}
\centering
\footnotesize 
\setlength{\tabcolsep}{8pt}
\renewcommand{\arraystretch}{1.1}
\begin{tabular}{lcccc}
\toprule
\textbf{Method} & \textbf{Acc} $\uparrow$ & \textbf{Pr} $\uparrow$ & \textbf{Re} $\uparrow$ & \textbf{F1} $\uparrow$ \\
\midrule
Ours L DGSA + MNN & 0.660 & 0.523 & 0.753 & 0.597 \\
\textcolor{blue}{\scriptsize (+FT gain)} 
                 & \textcolor{blue}{\scriptsize +0.032}
                 & \textcolor{blue}{\scriptsize +0.033}
                 & \textcolor{blue}{\scriptsize +0.057}
                 & \textcolor{blue}{\scriptsize +0.043} \\
\midrule
Ours H DGSA + MNN & 0.705 & 0.571 & 0.760 & 0.633 \\
\textcolor{blue}{\scriptsize (+FT gain)} 
                 & \textcolor{blue}{\scriptsize +0.024}
                 & \textcolor{blue}{\scriptsize +0.024}
                 & \textcolor{blue}{\scriptsize +0.076}
                 & \textcolor{blue}{\scriptsize +0.044} \\
\bottomrule
\end{tabular}

\end{table}

\subsubsection{Results with Model Trained with All Training Data}  \label{App: all data test}

In this experiment, we train our models using all available training data, including F2S data annotated with ScanNet labels, F2S data annotated with GPT-4o labels, and S2S data. The results are shown in Table~\ref{tab:all_data_results}. The gain is computed relative to the corresponding results reported in Table~\ref{tab: results simple}, \ref{tab: results medium}, \ref{tab: results difficult}, and \ref{tab: s2s results}.
For the S2S task, training with all data improves all metrics for both models. In particular, \textit{Ours L DGSA+MNN} improves F1 score by 4.2\%, while \textit{Ours H DGSA+MNN} achieves a larger gain of 5.6\%, with recall improving by up to 7.1\%. This indicates that additional training data enhances the model’s ability to identify correct correspondences in cross-scan matching.
For the F2S task, training with all data improves performance on the difficult test group annotated with GPT-4o labels, with F1 score increasing by 2.1\% and 1.9\% for \textit{Ours L DGSA+MCF} and \textit{Ours H DGSA+MCF}, respectively. In contrast, a slight decrease is observed on the simple and medium groups annotated with ScanNet labels. This is likely due to the increased label diversity introduced by GPT-4o annotations, which improves open-set generalization but slightly reduces performance on the more constrained ScanNet label set.
Overall, training with all available data improves performance in more challenging open-set and cross-scan scenarios, particularly for the S2S task and the difficult F2S group.

\begin{table}[t]
\centering
\footnotesize 
\caption{Results using the Model Trained with All Training Data} 
\label{tab:all_data_results}
\setlength{\tabcolsep}{3.5pt}
\renewcommand{\arraystretch}{1.15}
\begin{tabular}{clcccc}
\toprule
\textbf{Method} & \textbf{Task} & \textbf{Acc $\uparrow$} & \textbf{Pr $\uparrow$} & \textbf{Re $\uparrow$} & \textbf{F1 $\uparrow$} \\
\midrule
\multirow{6}{*}{Ours L DGSA+MCF} & F2S (Simple) & 0.826 & 0.826 & 0.989 & 0.884 \\
& \textcolor{blue}{\scriptsize (Gain)} 
                 & \textcolor{red}{\scriptsize -0.012}
                 & \textcolor{red}{\scriptsize -0.012}
                 & \textcolor{red}{\scriptsize -0.001}
                 & \textcolor{red}{\scriptsize -0.008} \\
& F2S (Medium) & 0.621 & 0.621 & 0.925 & 0.719 \\
& \textcolor{blue}{\scriptsize (Gain)} 
                 & \textcolor{red}{\scriptsize -0.005}
                 & \textcolor{red}{\scriptsize -0.005}
                 & {\scriptsize +0.000}
                 & \textcolor{red}{\scriptsize -0.004} \\
& F2S (Diffcult) & 0.376 & 0.376 & 0.806 & 0.487 \\
& \textcolor{blue}{\scriptsize (Gain)} 
                 & \textcolor{blue}{\scriptsize +0.020}
                 & \textcolor{blue}{\scriptsize +0.020}
                 & \textcolor{blue}{\scriptsize +0.018}
                 & \textcolor{blue}{\scriptsize +0.021} \\
\midrule
\multirow{2}{*}{Ours L DGSA+MNN} & S2S  & 0.662 & 0.527 & 0.742 & 0.596 \\
& \textcolor{blue}{\scriptsize (Gain)} 
                 & \textcolor{blue}{\scriptsize +0.034}
                 & \textcolor{blue}{\scriptsize +0.037}
                 & \textcolor{blue}{\scriptsize +0.046}
                 & \textcolor{blue}{\scriptsize +0.042} \\
\midrule
\multirow{6}{*}{Ours H DGSA+MCF} & F2S (Simple)& 0.858 & 0.884 & 0.937 & 0.897 \\
& \textcolor{blue}{\scriptsize (Gain)} 
                 & \textcolor{red}{\scriptsize -0.041}
                 & \textcolor{red}{\scriptsize -0.020}
                 & \textcolor{red}{\scriptsize -0.037}
                 & \textcolor{red}{\scriptsize -0.032} \\
& F2S (Medium) & 0.688 & 0.695 & 0.840 & 0.738 \\
& \textcolor{blue}{\scriptsize (Gain)} 
                 & \textcolor{red}{\scriptsize -0.026}
                 & \textcolor{red}{\scriptsize -0.014}
                 & \textcolor{red}{\scriptsize -0.047}
                 & \textcolor{red}{\scriptsize -0.030} \\
& F2S (Difficult) & 0.447 & 0.480 & 0.644 & 0.522 \\
& \textcolor{blue}{\scriptsize (Gain)} 
                 & \textcolor{blue}{\scriptsize +0.027}
                 & \textcolor{blue}{\scriptsize +0.040}
                 & \textcolor{red}{\scriptsize -0.027}
                 & \textcolor{blue}{\scriptsize +0.019} \\
\midrule
\multirow{2}{*}{Ours H DGSA+MNN} & S2S &  0.718 & 0.592 & 0.755 & 0.645 \\
& \textcolor{blue}{\scriptsize (Gain)} 
                 & \textcolor{blue}{\scriptsize +0.037}
                 & \textcolor{blue}{\scriptsize +0.045}
                 & \textcolor{blue}{\scriptsize +0.071}
                 & \textcolor{blue}{\scriptsize +0.056} \\
\bottomrule
\end{tabular}
\end{table}

\subsection{Downstream Task: Point Cloud Registration Result} \label{App: downstream Relocalization}
Scene graph alignment can improve downstream point cloud registration by providing high-level object correspondences, as demonstrated in several prior works (\cite{sarkar2023sgaligner,11024207}). Since the point clouds in our dataset contain instance-level information, scene graph alignment offers reliable object-level associations that can guide and constrain point-wise registration.
In this section, we evaluate whether scene graph alignment improves point cloud registration performance in both the F2S and S2S groups. In the F2S group, the task is to register a point cloud captured from the current frame to the point cloud of the full scan. We evaluate registration performance using correspondences predicted by our \textit{Ours H DGSA + MCF} model trained in Section~\ref{section: F2S results}. In the S2S group, the task is to register point clouds between two subscans, where we use correspondences predicted by our \textit{Ours H DGSA + MNN} model trained in Section~\ref{section: S2S results}.

\begin{table*}[t]
\caption{F2S Point Cloud Registration Results under Different RTE and RRE Thresholds (GT mIOU: 0.083)}
\label{tab:registration_results_f2s}
\centering
\footnotesize 
\setlength{\tabcolsep}{2.5pt}
\renewcommand{\arraystretch}{1.15}
\begin{tabular}{lcccccccccc}
\toprule
\multirow{2}{*}{\textbf{Correspondence Method}} 
& \multicolumn{3}{c}{\textbf{RTE}$\leq$\textbf{0.5m}, \textbf{RRE}$\leq$\textbf{5}$^\circ$} 
& \multicolumn{3}{c}{\textbf{RTE}$\leq$\textbf{1.0m}, \textbf{RRE}$\leq$\textbf{10}$^\circ$} 
& \multicolumn{3}{c}{\textbf{RTE}$\leq$\textbf{2.0m}, \textbf{RRE}$\leq$\textbf{15}$^\circ$} 
& \multirow{2}{*}{\textbf{Time$\downarrow$ (s)}} \\

& \textbf{SR$\uparrow$} 
& \textbf{RRE$\downarrow$ ($^\circ$) } 
& \textbf{RTE$\downarrow$ (m)} 
& \textbf{SR$\uparrow$} 
& \textbf{RRE$\downarrow$ ($^\circ$)} 
& \textbf{RTE$\downarrow$ (m)} 
& \textbf{SR$\uparrow$} 
& \textbf{RRE$\downarrow$ ($^\circ$)} 
& \textbf{RTE$\downarrow$ (m)} 
&  \\

\midrule
Global Points        
& 0.001 & 3.22  & 0.094 
& 0.002 & 6.55  & 0.365 
& 0.004 & 8.27 & 0.583 
& 4.55 \\

Global FPFH          
& \cellcolor{second}0.129 & \cellcolor{best}\textbf{1.91}  & \cellcolor{best}\textbf{0.076} 
& \cellcolor{second}0.148 & \cellcolor{best}\textbf{2.52}  & \cellcolor{best}\textbf{0.107} 
& 0.159 & \cellcolor{second}3.07  & 0.165 
& 4.48 \\

Our Matched Instances + One2One GeoTrans 
& 0.016 & 3.52  & 0.165 
& 0.057 & 6.41  & 0.231 
& 0.095 & 8.84  & 0.303 
& 2.73 \\

Our Matched Instances + One2One FPFH     
& 0.029 & 3.52  & 0.163 
& 0.107 & 6.50  & 0.224 
& \cellcolor{second}0.188 & 9.01  & 0.288 
& \cellcolor{second}1.50 \\

Our Matched Instances + Cropped FPFH
& \cellcolor{best}\textbf{0.201} & \cellcolor{second}1.97  &  \cellcolor{second}0.078 
& \cellcolor{best}\textbf{0.227} & \cellcolor{best}\textbf{2.52}  & \cellcolor{best}\textbf{0.107} 
& \cellcolor{best}\textbf{0.242} & \cellcolor{best}\textbf{3.00}  & \cellcolor{best}\textbf{0.163} 
& \cellcolor{best}\textbf{0.52} \\
\bottomrule
\end{tabular}
\end{table*}

\begin{table*}[t]
\caption{S2S Point Cloud Registration Results under Different RTE and RRE Thresholds (GT mIOU: 0.240)}
\label{tab:registration_results_s2s}
\centering
\footnotesize 
\setlength{\tabcolsep}{2.5pt}
\renewcommand{\arraystretch}{1.15}
\begin{tabular}{lcccccccccc}
\toprule
\multirow{2}{*}{\textbf{Correspondence Method}} 
& \multicolumn{3}{c}{\textbf{RTE}$\leq$\textbf{0.5m}, \textbf{RRE}$\leq$\textbf{5}$^\circ$} 
& \multicolumn{3}{c}{\textbf{RTE}$\leq$\textbf{1.0m}, \textbf{RRE}$\leq$\textbf{10}$^\circ$} 
& \multicolumn{3}{c}{\textbf{RTE}$\leq$\textbf{2.0m}, \textbf{RRE}$\leq$\textbf{15}$^\circ$} 
& \multirow{2}{*}{\textbf{Time$\downarrow$ (s)}} \\

& \textbf{SR$\uparrow$} 
& \textbf{RRE$\downarrow$ ($^\circ$) } 
& \textbf{RTE$\downarrow$ (m)} 
& \textbf{SR$\uparrow$} 
& \textbf{RRE$\downarrow$ ($^\circ$)} 
& \textbf{RTE$\downarrow$ (m)} 
& \textbf{SR$\uparrow$} 
& \textbf{RRE$\downarrow$ ($^\circ$)} 
& \textbf{RTE$\downarrow$ (m)} 
&  \\

\midrule
Global Points        
& 0.184 & \cellcolor{best}\textbf{2.02}  & 0.164 
& 0.246 & 3.10  & 0.292 
& 0.305 & 4.16 & 0.492 
& \cellcolor{best}\textbf{1.23} \\

Global FPFH          
& \cellcolor{best}\textbf{0.301} & 2.12  & \cellcolor{best}\textbf{0.145} 
& \cellcolor{second}0.372 & \cellcolor{second}2.78  & \cellcolor{second}0.222
& \cellcolor{second}0.402 & \cellcolor{best}\textbf{3.36}  & \cellcolor{best}\textbf{0.281} 
& 3.43 \\

Our Matched Instances + One2One GeoTrans
& 0.125 & 2.96  & 0.204 
& 0.181 & 4.07  & 0.267 
& 0.204 & 4.94  & 0.328 
& 2.04 \\

Our Matched Instances + One2One FPFH     
& 0.261 & 3.00  &  0.222 
& \cellcolor{best}\textbf{0.400} & 4.24  & 0.297 
& \cellcolor{best}\textbf{0.446} & 5.03  & 0.361 
& \cellcolor{second}1.71 \\

Our Matched Instances + Cropped FPFH     
& \cellcolor{second}0.297 & \cellcolor{second}2.04  &  \cellcolor{second}0.158 
& 0.355 & \cellcolor{best}\textbf{2.70}  & \cellcolor{best}\textbf{0.217} 
& 0.391 & \cellcolor{second}3.41  & \cellcolor{second}0.285 
& 2.40 \\

\bottomrule
\end{tabular}
\end{table*}

We conduct evaluation on test Scenes 600 to 625, resulting in 15,596 samples for the F2S group (using \textit{Cross Scan} data) and 708 samples for the S2S group. Registration in both groups is challenging due to the limited overlap between the source and target point clouds. In the F2S group, the frame point cloud represents only a small portion of the full scan. As a result, the overlap ratio is 0.891 when measured from frame to scan, but only 0.118 in the reverse direction (highly asymmetric overlap). The voxel-level (0.1,m resolution) mIoU between the two point clouds is only 0.083, indicating very limited geometric overlap. In the S2S group, the overlap ratios between subscans are 0.559 and 0.480 in the forward and reverse directions, respectively, with a voxel-level mIoU of 0.240.

We compare two baseline methods that do not use instance-level matching and three methods that leverage instance correspondences predicted by our model.
\begin{enumerate}[label=\textbf{\arabic*)}, leftmargin=*, itemsep=2pt]
\item  \textit{Global Points}: Point-wise correspondences are established directly on the full point clouds using RANSAC without feature extraction.
\item  \textit{Global FPFH}: Fast Point Feature Histogram (FPFH) features are first extracted to establish point correspondences, followed by RANSAC-based transformation estimation.
\item  \textit{Our Matched Instances + One2One GeoTransformer}: Inspired by SGAligner (\cite{sarkar2023sgaligner}), our model first predicts instance correspondences. GeoTransformer (\cite{qin2022geometric}) is then applied independently to each matched instance pair to establish point correspondences, which are aggregated and used in RANSAC for transformation estimation.
\item  \textit{Our Matched Instances + One2One FPFH}: This variant replaces GeoTransformer with FPFH features, which are extracted independently for each matched instance pair to establish correspondences prior to RANSAC.
\item \textit{Our Matched Instances + Cropped FPFH}: Our predicted instance correspondences are used to identify a spatial region enclosing the matched instances. A bounding box covering this region is computed, and only the cropped points within this region are used to extract FPFH features and establish correspondences for RANSAC. This approach leverages multiple matched instances jointly while excluding unrelated regions.
\end{enumerate}
For all methods, the estimated transformation is further refined using Iterative Closest Point (ICP) for fine registration. Note that the instance correspondences predicted by our model are not error-free, with detailed matching performance reported in Tables~\ref{tab: results medium} and \ref{tab: s2s results}.


Tables~\ref{tab:registration_results_f2s} and \ref{tab:registration_results_s2s} present the point cloud registration results for the F2S and S2S groups under different RTE and RRE thresholds.
As shown in Table~\ref{tab:registration_results_f2s}, global registration methods that operate on the entire point cloud without instance guidance perform poorly. In particular, \textit{Global Points} achieves almost zero success rate (SR), indicating that purely geometric point-wise RANSAC fails under such low overlap conditions. \textit{Global FPFH} improves performance by leveraging local geometric features, achieving SR of 0.129, 0.148, and 0.159 under increasingly relaxed thresholds, but remains fundamentally limited by the large proportion of irrelevant points in the full scan.
In contrast, methods leveraging instance-level correspondences significantly improve registration performance. Among them, \textit{Our Matched Instances + Cropped FPFH} achieves the best results across all thresholds, with SR improving to 0.201, 0.227, and 0.242.
This improvement demonstrates that scene graph alignment provides reliable object-level guidance to constrain correspondence search to semantically consistent regions, effectively mitigating the negative impact of low overlap. 
Furthermore, the cropped strategy reduces the number of outlier correspondences introduced by unrelated regions, resulting in both higher accuracy and faster computation. Specifically, \textit{Our Matched Instances + Cropped FPFH} achieves the lowest runtime (0.52 s), which is approximately 8.6× faster than global FPFH, highlighting the efficiency benefits of instance-guided spatial filtering. 
The comparison between \textit{One2One GeoTransformer}, \textit{One2One FPFH}, and \textit{Cropped FPFH} further reveals the importance of using multiple matched instances jointly. While One2One methods restrict correspondence estimation to individual instance pairs, the cropped approach leverages a larger but still relevant region containing multiple matched instances, providing more stable geometric constraints and improving robustness.

The S2S task is less difficult due to higher overlap between subscans. As shown in Table~\ref{tab:registration_results_s2s}, global methods already achieve reasonable performance. \textit{Global FPFH} achieves strong results across thresholds, reflecting that geometric features are sufficiently discriminative when overlap is moderate.
Instance-guided methods still provide clear advantages. In particular, \textit{Our Matched Instances + One2One FPFH} achieves the highest SR under moderate and relaxed thresholds, reaching 0.400 and 0.446, corresponding to relative improvements of 7.5\% and 10.9\% over \textit{Global FPFH}. This demonstrates that instance-level correspondences help eliminate geometrically ambiguous regions and guide registration toward semantically consistent structures. Meanwhile, \textit{Our Matched Instances + Cropped FPFH} achieves the lowest rotation and translation errors under moderate thresholds, indicating improved registration accuracy and stability.

Compared to the F2S task, the performance gap between global and instance-guided methods is smaller in S2S, as the higher geometric overlap makes global geometric features more reliable. 
Overall, these results demonstrate that our scene graph alignment model, despite containing imperfect instance correspondences, can significantly improve downstream point cloud registration, particularly in low-overlap scenarios. 
Instance-level correspondences constrain the correspondence search space to semantically and spatially relevant regions, reducing the likelihood of outlier matches and improving both efficiency and robustness.

\bibliographystyle{SageH}
\bibliography{head}

\end{document}